\documentclass[lettersize,journal]{IEEEtran}
\usepackage{algorithmic}
\usepackage{algorithm}
\usepackage{array}
\usepackage[caption=false,font=scriptsize,labelfont=sf,textfont=sf]{subfig}
\usepackage{textcomp}
\usepackage{stfloats}
\usepackage{url}
\usepackage{verbatim}
\usepackage{graphicx}
\usepackage{cite}
\usepackage[utf8]{inputenc} 
\usepackage[T1]{fontenc}    
\usepackage{hyperref}       
\usepackage{nicefrac}       
\usepackage{microtype}      
\usepackage{pifont}
\usepackage{multirow}   
\usepackage{colortbl}
\usepackage{amsmath, amssymb, amsfonts} 
\usepackage{booktabs}                   
\usepackage[table,xcdraw]{xcolor}       
\usepackage{makecell}
\usepackage{orcidlink}

\newtheorem{theorem}{Theorem}

\newtheorem{proposition}[theorem]{Proposition}

\begin{document}

\title{PRISM: Topology-Aware Cross-Modal Imputation for Modality-Deficient Federated Graph Learning}

\author{Zekai Chen, Miao Zhang, Jiayang Xing, Xunkai Li, Xun Wu, Rong-Hua Li\orcidlink{0000-0001-8658-6599}, Guoren Wang
\thanks{Zekai Chen, Miao Zhang, Jiayang Xing, Xunkai Li, Xun Wu, Rong-Hua Li, Guoren Wang are with Beijing Institute of Technology, Beijing, 100081, China.(e-mail:zackchen02@163.com;
1120243383@bit.edu.cn;cs2024xjy@bit.edu.cn;cs.xunkai.li@gmail.com;
alicewu0624@gmail.com;lironghuabit@126.com; wanggrbit@gmail.com)}
}

\maketitle

\begin{abstract}
Multimodal federated graph learning (MM-FGL) aims to collaboratively learn from decentralized graphs with text and images. However, real-world clients may not share a common modality basis: a visual-search client may contain image--interaction graphs but no seller descriptions, while a catalog client may provide text but no product images. We refer to this practical setting as client-level modality deficiency. Unlike random instance-wise missingness, a deficient client lacks the local semantic basis needed to reconstruct the absent modality. More importantly, in graph learning, incomplete representations initialize message passing, so imputation errors can be filtered, mixed, and amplified by the receiving topology. 
To address this gap, we propose \textbf{PRISM} (\textbf{P}roactive \textbf{R}etrieval and \textbf{I}mputation via \textbf{S}tructural \textbf{M}eta-prompting), a topology-aware federated cross-modal imputation framework. Rather than reconstructing the missing modality solely from local observations, PRISM recovers missing-modality semantics from the federation and introduces them into local graph propagation under topology-aware control. Experiments on six multimodal graph datasets across graph-centric and modality-centric tasks show that PRISM consistently improves modality-deficient clients, outperforming state-of-the-art baselines by \textbf{4.48}\% on average.
\end{abstract}


\section{Introduction}
\label{sec:intro}

\IEEEPARstart{G}{raph}-structured data are increasingly generated and stored in modern communication systems, including smartphone-based mobile social networks~\cite{pietilainen2009mobiclique,miluzzo2008sensing}, IoT sensing networks with spatial-temporal correlations~\cite{dong2023gnniot,li2018dcrnn}, V2X cooperative perception systems~\cite{xu2022v2xvit,hu2022where2comm}, and wearable sensor networks for human activity understanding~\cite{wang2023mhagnn}. Many of these applications are naturally multimodal, where users, devices, items, images, text, tags, and interactions are jointly organized as decentralized multimodal graphs~\cite{yan2025graph,li2026mmopenfgl}. Centralized mining of such data can enhance intelligent communication services, but it is often impractical because raw graph data are distributed across mobile or edge clients and are constrained by privacy risks, limited bandwidth, and wireless transmission costs. Federated graph learning (FGL) has therefore attracted increasing attention as a secure and communication-efficient paradigm for collaborative GNN training without exposing private nodes, edges, or attributes~\cite{he2021fedgraphnn,li2024openfgl}. Multimodal federated graph learning (MM-FGL) further aims to exploit both graph structure and cross-modal semantics across distributed communication environments while preserving raw-data locality.

\begin{figure}[t]
    \centering
    \includegraphics[width=0.95\linewidth]{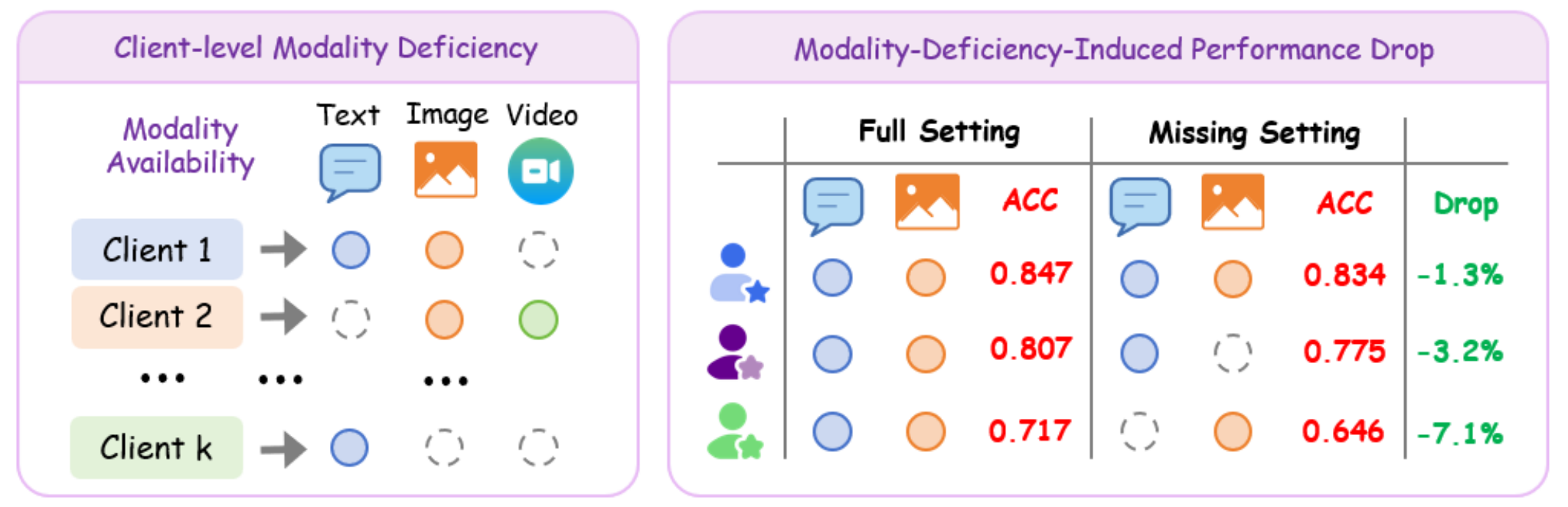}
    \vspace{-5pt}
    \caption{\textbf{Client-level modality deficiency and its federation-wide impact.}
    Left: clients may observe different modality channels, and some clients may permanently lack specific modalities.
    Right: a diagnostic example on the Toys node-classification task compares the same clients under full-modality and missing-modality settings.
    Missing modalities cause larger degradation on deficient clients, while modality-complete clients can also be affected through shared federated training.
    This indicates that client-level modality deficiency is a federation-wide challenge rather than a purely local missing-feature issue.}
    \label{fig:client_modality_deficiency}
\end{figure}

As illustrated in Fig.~\ref{fig:client_modality_deficiency}, real-world MM-FGL may violate the common-modality assumption implicitly adopted by most existing studies. 
Different mobile or edge clients may observe different modality channels due to heterogeneous sensing devices, platform-specific data-collection pipelines, licensing constraints, or intermittent modality availability. 
The diagnostic example in Fig.~\ref{fig:client_modality_deficiency} shows that this mismatch is not merely a local missing-feature issue. 
When some clients permanently lose an entire modality channel, their own performance drops substantially, and even modality-complete clients can be affected through shared federated training. 
This suggests that client-level modality deficiency can induce federation-wide performance degradation.

This modality deficiency introduces two coupled challenges. 
From the federated perspective, a deficient client lacks the local semantic basis of an entire modality and cannot recover it from its own observations alone. 
Moreover, because clients are coupled through shared model updates and cross-client semantic aggregation, modality-deficient clients may also affect the learned representation space of modality-complete clients. 
Since raw node attributes, missing-modality samples, and private edges cannot be exchanged across clients, the absent modality must be recovered only through privacy-preserving cross-client evidence. 
From the graph perspective, however, borrowing external semantics is risky. 
Unlike independent samples, where semantic incompleteness mainly affects the corresponding instance, node representations in a GNN serve as the initial conditions of neighborhood propagation. 
Once unreliable or incomplete representations enter the graph, their influence can be transmitted, mixed, and even amplified by local topology. 
In particular, an erroneous semantic cue attached to a structurally central node may affect a broad neighborhood after only a few propagation layers. 
We term this graph-specific failure mode a \emph{semantic boundary error}. 
In Sec.~\ref{sec:empirical_study}, we verify this phenomenon through empirical study.

These two challenges also explain why existing paradigms remain insufficient. 
On the one hand, multimodal graph completion and masked graph reconstruction are \emph{structurally aware but semantically under-sourced}~\cite{hou2022graphmae,jia2023multimodal,luanyuan2024mgnet}. 
They can exploit local topology to infer hidden information already supported by the observed feature space, but they cannot synthesize a modality basis that never appears locally. 
On the other hand, federated missing-modality learning is \emph{semantically informed but structurally uncontrolled}~\cite{tan2022fedproto,nguyen2024fedmac,singha2025fedmvp,xie2024mh}. 
Prototype sharing, cross-modal aggregation, and prompt-based adaptation can transfer useful semantics across clients, but they typically treat imputation as an instance-level supplement and do not control how the recovered semantics propagate through graph topology. 
Robust MM-FGL therefore requires a unified mechanism that can both obtain missing-modality semantics and regulate their topology-dependent influence in the receiving graph.

How can a graph client borrow a missing modality from the federation without propagating the wrong semantics through its topology? We answer this question with \textbf{PRISM} (\textbf{P}roactive \textbf{R}etrieval and \textbf{I}mputation via \textbf{S}tructural \textbf{M}eta-prompting), a topology-aware federated cross-modal imputation framework built on the principle of retrieve globally, inject structurally. PRISM separates \emph{where missing semantics come from} from \emph{how they enter graph propagation}. Specifically, it obtains missing-modality semantics from cross-client multimodal evidence, converts them into graph-compatible auxiliary signals, and adaptively regulates their influence according to the receiving client's topology and retrieval reliability. In this way, PRISM supplies the missing semantic basis while preventing unreliable semantics from being propagated uncontrollably during local message passing.

\textbf{Our Contributions.}
\underline{\textbf{\textit{(1) Problem Identification.}}}
We identify client-level modality deficiency as a distinct MM-FGL setting, where a client permanently lacks an entire modality. We further empirically reveal its graph-specific consequence, termed \emph{semantic boundary error}.
\underline{\textbf{\textit{(2) New Framework.}}}
We propose \textbf{PRISM}, a federated cross-modal imputation framework following the principle of retrieve globally, inject structurally. 
PRISM obtains missing-modality semantics from cross-client multimodal evidence, converts them into graph-compatible auxiliary signals, and regulates their influence according to the receiving client's topology and retrieval reliability. 
\underline{\textbf{\textit{(3) SOTA Performance.}}}
Experiments on six multimodal graph datasets across three tasks show that PRISM improves performance by \textbf{12.24}\% under missing-modality task settings, demonstrating its robustness under client-level modality deficiency.
\section{Preliminaries and Related Work}
\label{sec:prelim}

\subsection{Problem Formulation}
\label{subsec:problem}

\textbf{Multimodal Federated Graph Learning.}
We consider a multimodal federated graph learning (MM-FGL) system with $K$ clients. 
Client $k$ privately holds a subgraph 
$\mathcal{G}_k=(\mathcal{V}_k,\mathcal{E}_k,\mathbf{A}_k)$, 
where $n_k=|\mathcal{V}_k|$ and $\mathbf{A}_k$ is the local adjacency matrix. 
Let $\mathcal{M}$ denote the global modality set, such as image and text. 
In the full-modality setting, every node $v\in\mathcal{V}_k$ is associated with features $\{x_v^{(m)}\}_{m\in\mathcal{M}}$. 
Under \emph{client-level modality deficiency}, client $k$ observes only $\mathcal{M}_k\subseteq\mathcal{M}$ and lacks access to $\bar{\mathcal{M}}_k=\mathcal{M}\setminus\mathcal{M}_k$. 
Unlike random entry-wise missingness, this deficiency removes an entire modality basis from the client's local training signal.

Given the available modalities, a local multimodal encoder produces node representations
\begin{equation}
\mathbf{x}_v^{(0)}
=
f_k\!\left(\{x_v^{(m)}\}_{m\in\mathcal{M}_k}\right)\in\mathbb{R}^{d},
\quad
\mathbf{X}_k^{(0)}
=
\left[\mathbf{x}_v^{(0)}\right]_{v\in\mathcal{V}_k}.
\label{eq:node_modalities_prism}
\end{equation}
A GNN then propagates these features over the private graph. 
We first define the normalized propagation matrix as
\begin{equation}
\tilde{\mathbf{A}}_k
=
\mathbf{D}_k^{-1/2}
(\mathbf{A}_k+\mathbf{I})
\mathbf{D}_k^{-1/2}.
\label{eq:normalized_adj_prism}
\end{equation}
The layer-wise message-passing update is
\begin{equation}
\mathbf{H}_k^{(\ell+1)}
=
\sigma\!\left(
\tilde{\mathbf{A}}_k
\mathbf{H}_k^{(\ell)}
\mathbf{W}_{\mathrm{nbr}}^{(\ell)}
+
\mathbf{H}_k^{(\ell)}
\mathbf{W}_{\mathrm{self}}^{(\ell)}
\right).
\label{eq:gnn_mp_prism}
\end{equation}
The learning objective is to optimize shared parameters $\Theta$ and client-adaptive components $\{\Omega_k\}_{k=1}^{K}$ without exchanging raw node attributes, missing-modality samples, or private edges:
\begin{equation}
\min_{\Theta,\{\Omega_k\}_{k=1}^{K}}
\sum_{k=1}^{K}\frac{n_k}{\sum_{j=1}^{K}n_j}
\mathcal{L}_k\!\left(
g_{\Theta,\Omega_k}(\mathcal{G}_k,\{x_v^{(m)}\}_{m\in\mathcal{M}_k})
\right).
\label{eq:mmfgl_objective}
\end{equation}

\textbf{MM-FGL under Modality Deficiency.}
Under client-level modality deficiency, different clients participate in the same MM-FGL objective with different observable modality subsets. 
A full-modality client initializes node representations from the complete modality set $\mathcal{M}$, whereas a deficient client can only use $\mathcal{M}_k$ and permanently lacks the semantic basis in $\bar{\mathcal{M}}_k$. 
This setting is different from instance-wise missingness, because the missing modality is unavailable at the client level rather than absent for only a subset of nodes. 
Thus, modality-deficient MM-FGL requires recovering useful missing-modality semantics from cross-client evidence while preserving raw-data locality and avoiding uncontrolled propagation through the local graph.

\subsection{Related Work}
\label{subsec:related}

\textbf{Multimodal Graph Learning.}
Multimodal graph learning integrates graph topology with multimodal node attributes.. 
Existing methods typically design modality-aware propagation, heterogeneous attention, multi-graph interaction, or multimodal fusion mechanisms to improve graph-centric and modality-centric tasks~\cite{jia2023multimodal,luanyuan2024mgnet,yan2025graph}. 
These approaches provide useful architectures for modeling modality--topology interactions, but they generally assume centralized access to the graph or, at minimum, that each learner possesses a comparable modality basis.

\textbf{Federated Graph Learning.}
Federated graph learning extends FL to decentralized graph data while preserving the locality of nodes, edges, and attributes~\cite{he2021fedgraphnn,li2024openfgl}. 
Existing studies have explored a broad range of graph-specific challenges, including structural heterogeneity, personalized collaboration, and communication-efficient knowledge transfer~\cite{zhang2021subgraph,zhang2024deep,chen2025rethinking,aliakbari2025subgraph,tan2025s2fgl}. 
However, most FGL methods treat node attributes as unimodal features. 
They therefore do not address modality-level deficiency, where a client entirely lacks one semantic channel and has no local basis from which that modality can be reconstructed.

\textbf{Federated Multimodal and Missing-modality Learning.}
Federated multimodal learning studies collaboration when clients own heterogeneous or partially observed modalities, using prototype sharing, cross-modal aggregation, prompt-based adaptation, or personalized bypasses~\cite{tan2022fedproto,nguyen2024fedmac,singha2025fedmvp,xie2024mh}. 
However, most methods treat recovered semantics as node-wise feature supplements, overlooking that once injected into a GNN, they may propagate and contaminate neighboring representations. 
PRISM addresses this missing interface by retrieving sparse cross-modal prototypes from the federation and injecting them as topology-regulated virtual anchors, rather than as unrestricted feature additions.
\section{EMPIRICAL STUDY}
\label{sec:empirical_study}

Before introducing PRISM, we first examine a fundamental question: what exactly fails when client-level modality deficiency enters graph propagation? Our central hypothesis is that, unlike instance-wise missing features, client-level modality deficiency induces a \textit{semantic boundary error}. A graph client starts message passing from an incomplete semantic basis, and the resulting error is filtered, redistributed, or amplified by the receiving topology. We investigate this mechanism through three progressive questions: 
\textbf{Q1}: Does the same imputation error spread differently across client topologies? 
\textbf{Q2}: Are structurally central nodes more hazardous entry points for incorrect external semantics? 
\textbf{Q3}: Can topology-conditioned gating suppress the contamination caused by ambiguous imputation?

\subsection{Topology-Dependent Error Spreading (Q1)}
\label{subsec:q1_spreading}
We first examine whether an identical modality-induced perturbation evolves differently across diverse graph structures. For each dataset (Toys, KU, and QB), we construct full-modality node representations and then remove one modality from designated deficient clients (simulating Image-only and Text-only settings). This yields an observed deficient representation $X_k^{(0)}$ and a full-modality reference $X_k^*$ for client $k$. We define the initial semantic error as $\Delta_k = X_k^{(0)} - X_k^*$, and quantify its propagated magnitude after $L$ linearized message-passing steps by:
\begin{equation}
    R_k^{(L)} = \frac{\|S_k^L \Delta_k\|_F}{\|\Delta_k\|_F}.
\end{equation}
where $S_k$ denotes the normalized propagation operator of client $k$. A larger $R_k^{(L)}$ indicates error amplification, while a smaller value reflects attenuation.

\begin{figure}[!t]
    \centering
    \includegraphics[width=\linewidth]{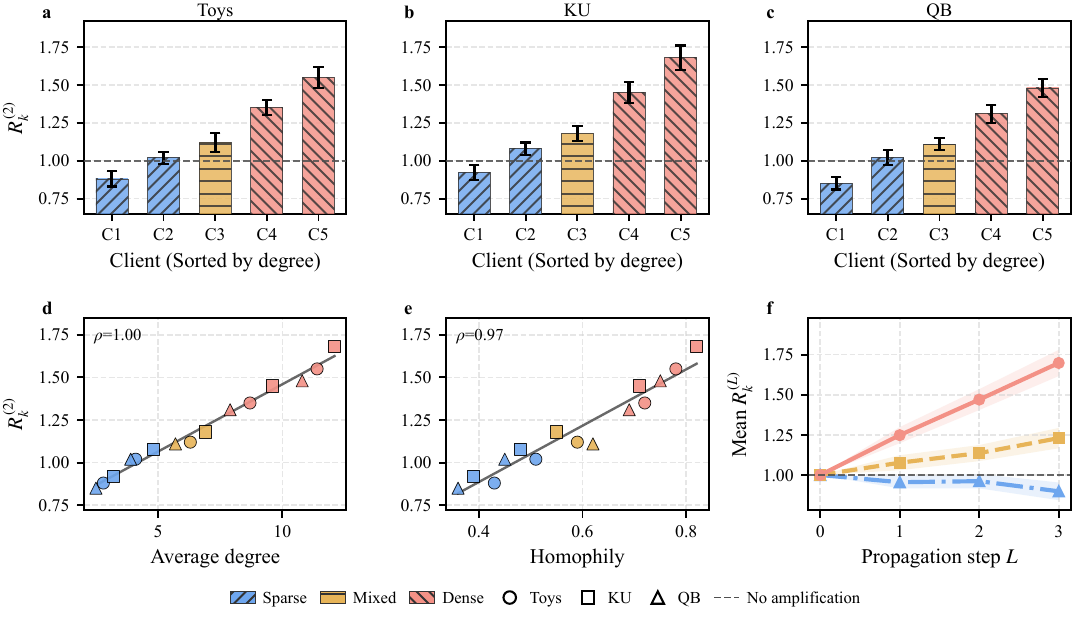}
    \vspace{-20pt}
    \caption{Topology-dependent error spreading. (a-c) Propagated error ratio $R_k^{(2)}$ across different client subgraphs on Toys, KU, and QB datasets. (d-f) Strong correlations between the propagated error ratio and structural statistics (Average Degree and Homophily), as well as the progression across propagation steps $L$.}
    \label{fig:topology_spreading}
\end{figure}

As shown in Fig. \ref{fig:topology_spreading}(a-c), across clients, the same modality-induced perturbation produces substantially different propagated error ratios. Dense or highly homophilous client graphs tend to preserve and diffuse the error more strongly, reaching approximately 1.42--1.58 times the initial error scale ($R_k^{(2)} > 1$). Conversely, sparse or structurally fragmented subgraphs show lower average ratios (around 0.91--1.07) but exhibit noticeably larger variance, as a few bridge-like edges can still transmit the perturbation. Furthermore, as shown in Fig. \ref{fig:topology_spreading}(d-e), $R_k^{(2)}$ is highly correlated with structural metrics such as average degree ($\rho=1.00$) and homophily ($\rho=0.98$). These results confirm that missing modalities are not merely local feature gaps; once propagation begins, their downstream impact becomes highly topology-dependent.

\subsection{Structure-Sensitive Injection Risk (Q2)}
\label{subsec:q2_injection}
We next ask whether the harm of incorrect semantic injection depends on where the error enters the graph. To isolate the effect of structural position, we inject a controlled semantic perturbation $\delta$ into selected nodes and measure the resulting representation shift on their $h$-hop neighborhoods. We define the $h$-hop neighbor contamination score for an injected node set $A$ as:
\begin{equation}
    \mathrm{Contam}^{(h)}(A) = \frac{1}{|\mathcal{N}_h(A)|} \sum_{u\in \mathcal{N}_h(A)} \frac{\|H_u^{\mathrm{pert}} - H_u^{\mathrm{clean}}\|_2}{\|\delta\|_2}.
\end{equation}
and compare the Top-20\% most central nodes (via PageRank) against the Bottom-20\% peripheral nodes. 

\begin{figure}[!t]
    \centering
    \includegraphics[width=\linewidth]{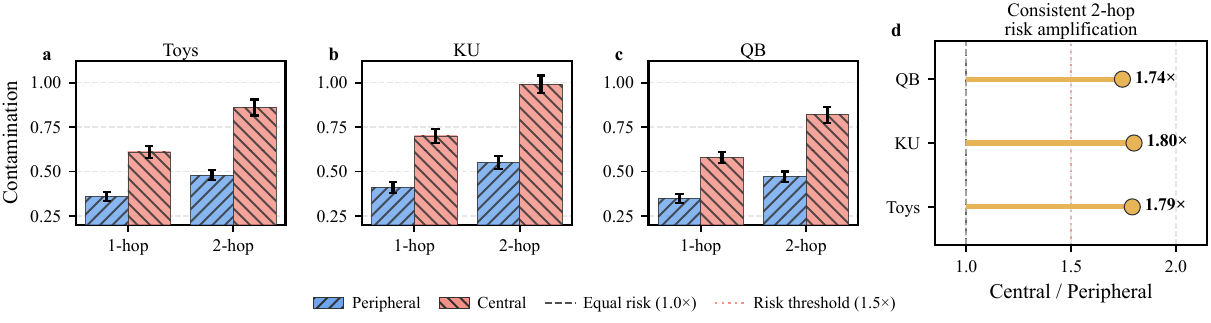}
    \vspace{-20pt}
    \caption{Central vs. Peripheral semantic injection risks. (a-c) Contamination scores for 1-hop and 2-hop neighborhoods when errors are injected into Top-20\% central versus Bottom-20\% peripheral nodes. (d) The 2-hop Central/Peripheral contamination ratio consistently exceeds 1.5$\times$ across datasets.}
    \label{fig:central_injection}
\end{figure}

As illustrated in Fig. \ref{fig:central_injection}, injecting the same wrong semantic vector into structurally central nodes causes markedly larger neighborhood contamination than injecting it into peripheral nodes. This effect becomes especially pronounced after two propagation steps, where errors introduced near hubs influence a much larger portion of the local subgraph. The two-hop contamination ratio approaches 1.8$\times$ across datasets (Fig. \ref{fig:central_injection}d). This indicates that the risk of semantic imputation in MM-FGL depends not only on whether the retrieved semantics are accurate, but also on how structurally influential their injection locations are.

\subsection{Topology-Conditioned Gating Control (Q3)}
\label{subsec:q3_gating}
Finally, we examine whether controlling the propagation strength of retrieved semantics can mitigate the aforementioned risk. We divide server-side retrievals into low-margin (ambiguous) and high-margin (confident) groups, comparing fixed-strength injection against confidence-only and topology-conditioned gating strategies.

\begin{figure}[!t]
    \centering
    \includegraphics[width=\linewidth]{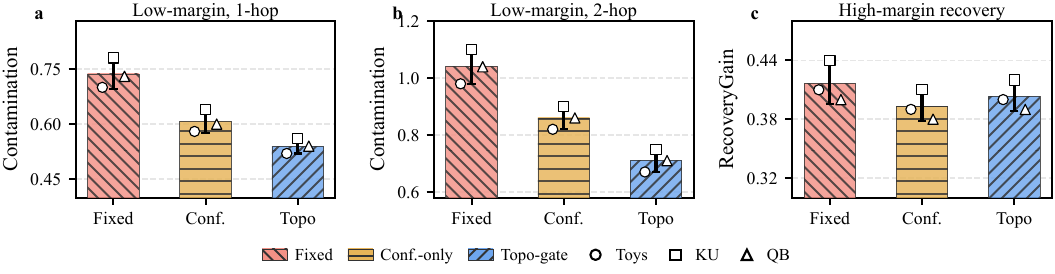}
    \vspace{-20pt}
    \caption{Effectiveness of topology-conditioned gating. (a-b) Under low-margin (ambiguous) retrieval, topology-conditioned gating significantly reduces 1-hop and 2-hop contamination compared to fixed injection. (c) Under high-margin retrieval, the gate preserves the recovery gain (RecoveryGain), ensuring semantic recovery without structural risk.}
    \label{fig:gating_control}
\end{figure}

As shown in Fig. \ref{fig:gating_control}, fixed injection severely amplifies contamination when retrieval confidence is low. By contrast, under low-margin retrievals, topology-conditioned gating substantially reduces one-hop and two-hop contamination (by 23.6\% and 31.4\%, respectively) compared with fixed-strength injection. Importantly, Fig. \ref{fig:gating_control}(c) demonstrates that the gate preserves most of the recovery benefit (measured by RecoveryGain) under high-margin retrievals. 

Taken together, these findings show that robust cross-modal imputation in federated graphs requires more than importing external semantics: it must also regulate their propagation budget according to both retrieval reliability and the structural properties of the receiving graph. This establishes the core design principle of PRISM: \textit{retrieve globally, inject structurally}.
\section{Methodology}
\label{sec:methodology}

\begin{figure*}[t]
  \centering
  \includegraphics[width=0.998\textwidth]{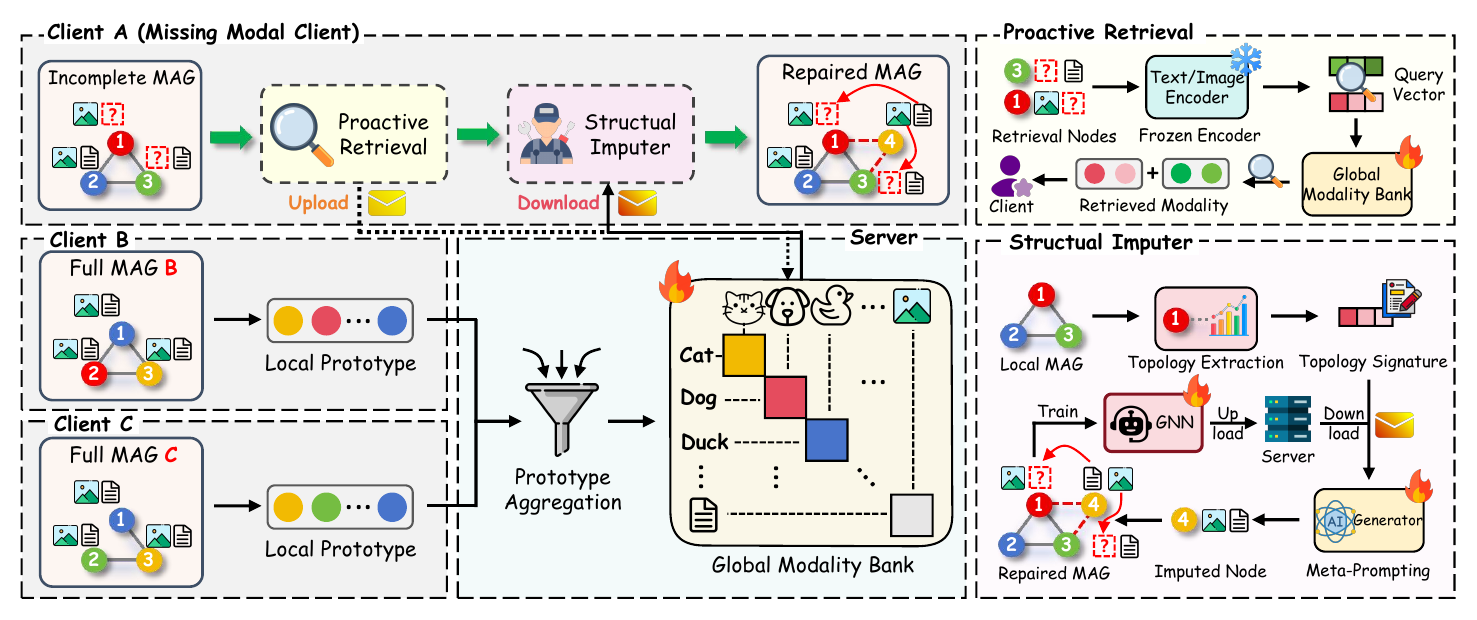}
  \vspace{-20pt}
  \caption{Overview of PRISM. A modality-deficient client summarizes its surviving-modality graph into semantic queries, retrieves sparse prototypes from the server-side modality bank, instantiates them as virtual anchors, and injects them through a topology-conditioned gate during local message passing.}
  \label{fig:method}
\end{figure*}

\subsection{Overview}
\label{sec:overview}

PRISM follows the principle \emph{retrieve globally, inject structurally}. As illustrated in Fig.~\ref{fig:method}, PRISM addresses these two requirements through three coupled components in each communication round: a server-side \emph{Global Modality Bank} that stores multimodal prototype keys and values, sparse prototype retrieval driven by the client's surviving modalities, and virtual-anchor injection into local message passing. A structural meta-prompt derived from lightweight spectral and motif statistics further adapts both the retrieval projection and the anchor-injection gate, enabling PRISM to borrow missing semantics while controlling their graph-level influence. We detail these components in Secs.~\ref{sec:global-bank}--\ref{sec:federated-optimization}.

\begin{algorithm}[t]
\caption{Overview of PRISM}
\label{alg:prism}
\begin{algorithmic}[1]
\REQUIRE Clients $\{\mathcal{G}_k\}_{k=1}^{K}$, global bank size $J$, warm-up rounds $T_w$.
\ENSURE Global parameters $\Theta,\Phi$ and modality bank $\mathcal{P}$.
\STATE Initialize $\Theta^{(0)},\Phi^{(0)}$ and $\mathcal{P}^{(0)}$ by Eq.~\eqref{eq:prototype_bank}.
\FOR{each communication round $t=0,\ldots,T-1$}
    \FOR{each client $k$ \textbf{in parallel}}
        \IF{$t<T_w$}
            \STATE Train local GNN without anchor injection.
        \ELSE
            \STATE Compute structural descriptor $\mathbf{z}_k$ and projection $\mathbf{P}_k$ by Eqs.~\eqref{eq:topo_signature}--\eqref{eq:prompt_projection}.
            \IF{client $k$ is modality-deficient}
                \STATE Form cluster queries by Eq.~\eqref{eq:cluster_query} and retrieve top-$s$ prototypes by Eq.~\eqref{eq:retrieval}.
                \STATE Construct virtual anchors and confidence gate by Eqs.~\eqref{eq:aug_graph}--\eqref{eq:gamma_gate}.
                \STATE Perform anchor-augmented message passing by Eq.~\eqref{eq:anchor_mp}.
            \ELSE
                \STATE Perform standard full-modality message passing.
            \ENDIF
        \ENDIF
        \STATE Optimize by Eqs.~\eqref{eq:total_loss}--\eqref{eq:meta_loss} and upload updates.
    \ENDFOR
    \STATE Server refreshes $\mathcal{P}$ by Eq.~\eqref{eq:bank_ema} and aggregates $\Theta,\Phi$ by Eq.~\eqref{eq:federated_update}.
\ENDFOR
\end{algorithmic}
\end{algorithm}

\subsection{Global Modality Bank}
\label{sec:global-bank}

\textbf{Motivation.}
A modality-deficient client cannot reconstruct an absent modality from local topology alone, because the corresponding semantic basis may never appear in its training data. 
PRISM therefore moves the source of missing semantics from local reconstruction to federated retrieval: the server maintains a compact global bank that accumulates cross-client multimodal evidence without exchanging raw nodes, edges, or modality samples.

\textbf{Federated semantic basis.}
We assume that at least a subset of clients observe paired multimodal evidence, allowing the server-side bank to form an aligned semantic basis. 
If the federation contains no evidence for a missing modality, PRISM cannot synthesize that modality from nothing and reduces to topology-aware regularization without cross-modal recovery. 
The server maintains $J$ multimodal prototypes
\begin{equation}
\mathcal{P}=\{p_j\}_{j=1}^{J}, \quad
p_j=\big(\{k_j^{(m)}\}_{m\in\mathcal{M}},\, p_j^{(v)}\big).
\label{eq:prototype_bank}
\end{equation}
where $k_j^{(m)}\in\mathbb{R}^{d_q}$ is the retrieval key for modality $m$, and $p_j^{(v)}\in\mathbb{R}^{d}$ is the prototype value used as semantic content. 
This key--value design separates semantic indexing from content transfer: compact keys support retrieval, while dense values are transmitted only for selected prototypes.

\textbf{Prototype update.}
At initialization, the server builds the bank from normalized cluster centroids uploaded by warm-up clients. 
In later rounds, each participating client forms candidate updates from clusters whose assignment confidence exceeds a threshold $\xi$. 
A deficient client only updates keys for modalities it observes, whereas a full-modal client contributes aligned multimodal key--value pairs. 
The server matches each candidate to its nearest prototype key by cosine similarity; unmatched high-confidence candidates refresh low-usage prototypes. 
Matched updates are aggregated by client-size-weighted averaging and smoothed with exponential moving average:
\begin{equation}
p_j^{(t+1)} \leftarrow \beta_{\mathrm{g}}p_j^{(t)}
+(1-\beta_{\mathrm{g}})\tilde{p}_j^{(t)}.
\label{eq:bank_ema}
\end{equation}
where $\tilde{p}_j^{(t)}$ is the aggregated prototype update at round $t$. 
Only prototype-level summaries are exchanged, preserving the standard raw-data locality boundary of federated learning. 
Clients do not transmit raw node attributes, missing-modality samples, or private edges. 
Prototype summaries may still leak aggregate information under strong adversaries; formal privacy guarantees can be incorporated through secure aggregation, norm clipping, or differential privacy, which are orthogonal to PRISM's retrieval-and-injection design.

\textbf{Training procedure.}
Each communication round follows four steps: 
(1) selected clients update local encoders and form confidence-filtered cluster summaries; 
(2) the server refreshes the modality bank through nearest-prototype matching and EMA smoothing; 
(3) modality-deficient clients retrieve top-$s$ prototypes using the topology-conditioned metric; and 
(4) clients perform anchor-augmented local training before federated averaging. 
The exchanged payload consists of cluster-level keys and selected prototype values rather than node-level representations.

\subsection{Proactive Cross-Modal Retrieval}
\label{sec:proactive-retrieval}

\textbf{Motivation.}
The global bank provides a federation-level modality basis, but a deficient client must still identify which external semantics are relevant to its own graph. 
Direct node-level retrieval is both noisy and communication-intensive. 
PRISM therefore performs cluster-level retrieval using the client's surviving modality, yielding compact and more stable semantic queries.

\textbf{Querying with the surviving modality.}
Client $k$ first computes stable node representations $\bar{\mathbf{h}}_v$ with a momentum encoder. 
To reduce retrieval noise and communication cost, nodes are grouped into $C$ local semantic clusters $\{\mathcal{C}_r\}_{r=1}^{C}$. 
For an available modality $m\in\mathcal{M}_k$, each cluster is summarized and projected into the retrieval space as
\begin{equation}
\mathbf{q}_{r}^{(m)}
=
\left(
\frac{1}{|\mathcal{C}_r|}
\sum_{v\in\mathcal{C}_r}
\bar{\mathbf{h}}_v
\right)
\mathbf{W}_{Q}^{(m)}.
\label{eq:cluster_query}
\end{equation}

\textbf{Sparse topology-conditioned retrieval.}
A topology-agnostic query would select the same semantic dimensions regardless of how the receiving graph propagates information. 
PRISM therefore transforms the query using a topology-conditioned projection $\mathbf{P}_k$ generated in Sec.~\ref{sec:structural-imputer}. 
The client retrieves only the top-$s$ prototypes:
\begin{equation}
\mathcal{I}_{r}
=
\operatorname{Top}_{s}
\left\{
\left\langle
\mathbf{q}_{r}^{(m)}\mathbf{P}_k,\,
k_j^{(m)}
\right\rangle
\right\}_{j=1}^{J}.
\label{eq:retrieval}
\end{equation}
The retrieved values $\{p_j^{(v)}:j\in\mathcal{I}_{r}\}$ serve as a compact semantic substitute for the missing modality. 
Sparsity is essential here: irrelevant prototypes are not harmless, since once injected into a GNN they may be propagated across neighborhoods.

\subsection{Structural Meta-Prompting}
\label{sec:structural-imputer}

\textbf{Motivation.}
The same external semantic cue may behave very differently across client graphs. 
Dense and homophilous graphs can often tolerate stronger semantic smoothing, whereas sparse, fragmented, or heterophilous graphs are more vulnerable to noisy anchor propagation. 
To prevent retrieved semantics from becoming topology-dependent boundary errors, PRISM conditions both retrieval and injection on a lightweight structural prior.

\textbf{Topology as an injection prior.}
For client $k$, we compute
\begin{equation}
\mathbf{z}_k
=
\operatorname{LayerNorm}
\left(
[\log\lambda_1,\ldots,\log\lambda_r,d_{\mathrm{tri}},d_{\mathrm{wedge}}]
\right).
\label{eq:topo_signature}
\end{equation}
where $\{\lambda_i\}_{i=1}^{r}$ are the leading non-trivial eigenvalues of the normalized graph Laplacian, and $d_{\mathrm{tri}}$ and $d_{\mathrm{wedge}}$ denote triangle and wedge densities. 
This descriptor captures both global spectral smoothness and local motif structure, providing a compact summary of how the receiving graph may propagate external semantics.

\textbf{Low-rank prompt basis.}
To avoid a large client-specific controller, PRISM learns a low-rank prompt basis. 
Here, ``meta-prompt'' denotes a topology-generated control vector that modulates retrieval and injection, rather than a text prompt. 
Let $\{\mathbf{u}_b\}_{b=1}^{B}$ be global basis vectors and $\mathbf{u}_0$ be a shared base prompt. 
The topology-conditioned retrieval projection is generated as
\begin{equation}
\boldsymbol{\omega}_k
=
\operatorname{softmax}
\left(
\frac{\mathbf{W}_{z}\mathbf{z}_k}{\tau}
\right),
\mathbf{P}_k
=
\operatorname{diag}
\left(
\mathbf{u}_0+
\sum_{b=1}^{B}
\omega_{k,b}\mathbf{u}_b
\right).
\label{eq:prompt_projection}
\end{equation}
The prompt reshapes the retrieval metric before external semantics enter the graph, deciding which semantic dimensions should be trusted under the current topology.

\subsection{Virtual Anchor Injection}
\label{sec:anchor-injection}

\textbf{Motivation.}
Retrieval determines \emph{what} semantics should be borrowed, but not \emph{how} they should enter graph propagation. 
Directly concatenating retrieved features to node representations ignores the graph-specific risk identified in Sec.~\ref{sec:empirical_study}: unreliable semantics may contaminate neighborhoods once message passing begins. 
PRISM instead introduces retrieved prototypes as virtual anchors and controls their propagation strength explicitly.

\textbf{From retrieved prototypes to graph messages.}
For each cluster $\mathcal{C}_r$, retrieved prototypes are instantiated as virtual nodes:
Let $\mathcal{A}_r=\{a_{r,j}\}_{j\in\mathcal{I}_r}$ denote the virtual anchors retrieved for cluster $\mathcal{C}_r$. The augmented graph is
\begin{equation}
\tilde{\mathcal{V}}_k
=
\mathcal{V}_k\cup\mathcal{A}_r,
\quad
\tilde{\mathcal{E}}_k
=
\mathcal{E}_k\cup
(\mathcal{A}_r\times\mathcal{C}_r).
\label{eq:aug_graph}
\end{equation}
Each anchor $a_{r,j}$ is initialized by $p_j^{(v)}$. 
To suppress ambiguous retrievals, PRISM computes a confidence gate from the retrieval margin:
\begin{equation}
\gamma_r
=
\operatorname{clip}
\left(
\sigma\!\left(a(S_{r,1}-S_{r,2}-c_0)\right),
0,\gamma_{\max}
\right).
\label{eq:gamma_gate}
\end{equation}
where $S_{r,1}$ and $S_{r,2}$ are the top-1 and top-2 retrieval scores. 
A small margin indicates that the bank does not provide a clear semantic match, and the corresponding anchor influence is therefore down-weighted.

The local GNN then performs anchor-augmented message passing:
\begin{equation}
\mathbf{h}_v^{(\ell+1)}
=
\operatorname{UPD}^{(\ell)}
\left(
\mathbf{h}_v^{(\ell)},
\sum_{u\in\mathcal{N}(v)}\alpha_{vu}\mathbf{h}_u^{(\ell)}
+
\gamma_r
\sum_{j\in\mathcal{I}_r}
\rho_{r,j}p_j^{(v)}
\right).
\label{eq:anchor_mp}
\end{equation}
where $v\in\mathcal{C}_r$, $\rho_{r,j}$ is the normalized retrieval weight, and $\alpha_{vu}$ is the original graph aggregation weight. 
Thus, missing semantics are injected as confidence-controlled graph messages rather than unrestricted feature additions.

\subsection{Federated Optimization}
\label{sec:federated-optimization}

\textbf{Motivation.}
PRISM introduces an additional retrieval-and-injection loop, whose reliability depends on stable local representations and a meaningful global bank. 
Activating cross-modal retrieval too early may reinforce noisy associations before the semantic space is sufficiently formed. 
We therefore adopt a two-phase optimization strategy that first stabilizes local representations and then jointly trains task prediction, retrieval alignment, and topology-aware control.

During warm-up, each client trains its local encoder and GNN backbone without anchor injection. 
After warm-up, clients activate PRISM and optimize
\begin{equation}
\mathcal{L}_{k}
=
\mathcal{L}_{\mathrm{task}}^{(k)}
+\lambda_{\mathrm{ret}}\mathcal{L}_{\mathrm{ret}}^{(k)}
+\lambda_{\mathrm{meta}}\mathcal{L}_{\mathrm{meta}}^{(k)}.
\label{eq:total_loss}
\end{equation}
Here $\mathcal{L}_{\mathrm{task}}^{(k)}$ is the downstream task loss. 
The retrieval loss aligns confident anchors with local cluster representations:
\begin{equation}
\mathcal{L}_{\mathrm{ret}}^{(k)}
=
-\frac{1}{C}\sum_{r=1}^{C}
\log
\frac{\exp(\operatorname{sim}(\mathbf{c}_r,\sum_{j\in\mathcal{I}_r}\rho_{r,j}p_j^{(v)})/\tau_r)}
{\sum_{j'=1}^{J}\exp(\operatorname{sim}(\mathbf{c}_r,p_{j'}^{(v)})/\tau_r)}.
\label{eq:retrieval_loss}
\end{equation}
while the meta-prompt regularizer discourages degenerate topology controllers:
\begin{equation}
\mathcal{L}_{\mathrm{meta}}^{(k)}
=
\|\mathbf{P}_k-\mathbf{I}\|_F^2
+\beta_{\omega}\sum_{b=1}^{B}\omega_{k,b}\log\omega_{k,b}.
\label{eq:meta_loss}
\end{equation}
The first term keeps the topology-conditioned metric close to the shared retrieval space unless structural evidence supports adaptation, while the entropy term prevents all clients from collapsing to the same basis element.

The server aggregates the global backbone and prompt parameters by weighted averaging:
\begin{equation}
\Theta^{(t+1)}
=
\sum_{k=1}^{K}
\frac{|\mathcal{V}_k|}{\sum_{k'}|\mathcal{V}_{k'}|}
\Theta_k^{(t)},
\Phi^{(t+1)}
=
\sum_{k=1}^{K}
\frac{|\mathcal{V}_k|}{\sum_{k'}|\mathcal{V}_{k'}|}
\Phi_k^{(t)}.
\label{eq:federated_update}
\end{equation}
where $\Phi=\{\mathbf{W}_z,\mathbf{u}_0,\{\mathbf{u}_b\}_{b=1}^{B}\}$ denotes the structural meta-prompt parameters. 
This optimization remains compatible with standard FGL pipelines while adding a retrieval-and-injection layer for modality-deficient clients. 
Overall, PRISM forms a closed semantic recovery loop: the global bank supplies the missing modality basis, structural prompting adapts retrieval to the receiving topology, and confidence-aware anchors constrain how external semantics propagate through the graph.

\section{Theoretical Analysis}
\label{sec:theoretical}

This section provides a mechanistic analysis of PRISM. 
Rather than establishing a full convergence or generalization guarantee, we analyze the three properties that directly support our design principle, \emph{retrieve globally, inject structurally}. 
First, client-level modality deficiency creates local semantic uncertainty that cannot be eliminated without cross-client semantic evidence. 
Second, recovered semantics may still induce topology-mediated boundary risk once injected into graph propagation. 
Third, PRISM reduces semantic communication by exchanging cluster-level summaries and sparse prototype values instead of node-level missing-modality features. 
All proofs are provided in the supplementary materials.

\subsection{Analysis Setup}

For client $k$ and semantic cluster $\mathcal{C}_r$, let 
$\mathbf{x}_{k,r}^{*}\in\mathbb{R}^{d}$ denote the unavailable full-modality semantic target, and let $\mathcal{O}_{k,r}$ denote all locally observable information, including the surviving modality and local topology. 
PRISM retrieves a sparse semantic substitute $\hat{\mathbf{x}}_{k,r}$ from the global modality bank and injects it into local graph propagation with strength $\gamma_{k,r}\geq 0$. 
Let $\mathbf{S}_k$ be the normalized propagation operator of client $k$, and let $\mathbf{b}_{k,r}$ be the normalized indicator vector of cluster $\mathcal{C}_r$.

\subsection{Local Semantic Uncertainty}

A modality-deficient client may never observe the target modality locally. 
Thus, the missing semantic basis cannot always be inferred from the surviving modalities and the local topology alone.

\begin{theorem}[Local Irrecoverability Bound]
\label{thm:local_irrecoverability}
For any local estimator $f_k(\mathcal{O}_{k,r})$ that does not access cross-client semantic evidence,
\begin{equation}
\mathbb{E}
\left[
\left\|
\mathbf{x}_{k,r}^{*}
-
f_k(\mathcal{O}_{k,r})
\right\|_2^2
\right]
\geq
\mathbb{E}
\left[
\operatorname{Tr}
\left(
\operatorname{Cov}
\left[
\mathbf{x}_{k,r}^{*}
\mid
\mathcal{O}_{k,r}
\right]
\right)
\right].
\label{eq:local_irrecoverability}
\end{equation}
\end{theorem}

The bound is positive whenever the absent modality is not conditionally determined by local observations. 
This does not imply that every missing modality is unrecoverable, but it formalizes a key limitation of client-level modality deficiency: if the missing semantic basis is not locally identifiable, then purely local completion suffers from irreducible uncertainty. 
This motivates the federation-level retrieval component of PRISM.

\subsection{Topology-mediated Boundary Risk}

Federation-level retrieval provides a semantic source, but it does not by itself determine how recovered semantics should enter graph propagation. 
If the retrieved substitute is imperfect, its residual error may spread through the receiving topology. 
For cluster $\mathcal{C}_r$, define the residual as
\begin{equation}
\mathbf{e}_{k,r}
=
\hat{\mathbf{x}}_{k,r}
-
\mathbf{x}_{k,r}^{*}.
\label{eq:semantic_residual}
\end{equation}
After anchor injection with strength $\gamma_{k,r}$, the residual entering message passing is
\begin{equation}
\mathbf{E}_{k,r}
=
\gamma_{k,r}
\mathbf{b}_{k,r}
\mathbf{e}_{k,r}^{\top}.
\label{eq:boundary_residual}
\end{equation}
We define the $L$-step semantic boundary risk as
\begin{equation}
\mathcal{R}_{k,r}^{(L)}
=
\left\|
\mathbf{S}_k^{L}
\mathbf{E}_{k,r}
\right\|_{F}.
\label{eq:semantic_boundary_risk}
\end{equation}

\begin{theorem}[Semantic Boundary Risk Factorization]
\label{thm:boundary_factorization}
The propagated semantic boundary risk satisfies
\begin{equation}
\mathcal{R}_{k,r}^{(L)}
=
\gamma_{k,r}
\chi_{k,r}^{(L)}
\left\|
\mathbf{e}_{k,r}
\right\|_2,
\quad
\chi_{k,r}^{(L)}
=
\left\|
\mathbf{S}_k^{L}
\mathbf{b}_{k,r}
\right\|_2.
\label{eq:boundary_factorization}
\end{equation}
If $\mathbf{S}_k=\mathbf{U}\boldsymbol{\Lambda}\mathbf{U}^{\top}$ is symmetric, then
\begin{equation}
\left(
\chi_{k,r}^{(L)}
\right)^2
=
\sum_i
\lambda_i^{2L}
\left(
\mathbf{u}_i^{\top}
\mathbf{b}_{k,r}
\right)^2.
\label{eq:structural_exposure_spectral}
\end{equation}
\end{theorem}

Theorem~\ref{thm:boundary_factorization} decomposes boundary risk into semantic residual magnitude, injection strength, and structural exposure. 
Hence, the same semantic residual may remain localized in one client graph but contaminate a wider neighborhood in another. 
This result formalizes the graph-specific failure mode behind semantic boundary errors and justifies the need for topology-aware injection.

\subsection{Sparse Retrieval and Confidence-controlled Injection}

PRISM controls the residual term in Theorem~\ref{thm:boundary_factorization} through sparse retrieval from the global modality bank. 
Let $\mathcal{P}=[\mathbf{p}_1,\ldots,\mathbf{p}_J]^\top$ be the prototype matrix with $\|\mathbf{p}_j\|_2\leq P_{\max}$. 
Let $\boldsymbol{\pi}_{k,r}(\mathbf{P}_k)$ denote the retrieved sparse distribution under the topology-conditioned metric $\mathbf{P}_k$, where $\|\boldsymbol{\pi}_{k,r}(\mathbf{P}_k)\|_0\leq s$, and
\begin{equation}
\hat{\mathbf{x}}_{k,r}
=
\mathcal{P}^{\top}
\boldsymbol{\pi}_{k,r}(\mathbf{P}_k).
\label{eq:retrieved_semantic_substitute}
\end{equation}
For support size $s$, define the best sparse approximation error as
\begin{equation}
\epsilon_s(\mathbf{x}_{k,r}^{*})
=
\min_{\boldsymbol{\pi}\in\Delta^{J-1},\,\|\boldsymbol{\pi}\|_0\leq s}
\left\|
\mathcal{P}^{\top}\boldsymbol{\pi}
-
\mathbf{x}_{k,r}^{*}
\right\|_2.
\label{eq:sparse_bank_error_theory}
\end{equation}
Let $\boldsymbol{\pi}_{k,r}^{*,s}$ be an optimal solution of Eq.~\eqref{eq:sparse_bank_error_theory}. 
The retrieval mismatch under metric $\mathbf{P}_k$ is
\begin{equation}
\begin{gathered}
\delta_s(\mathbf{P}_k)
=
\left\|
\left(
\boldsymbol{\pi}_{k,r}(\mathbf{P}_k)
-
\boldsymbol{\pi}_{k,r}^{*,s}
\right)_{\mathcal{S}_{k,r}}
\right\|_2, \\
\mathcal{S}_{k,r}
=
\operatorname{supp}
\left(
\boldsymbol{\pi}_{k,r}(\mathbf{P}_k)
\right)
\cup
\operatorname{supp}
\left(
\boldsymbol{\pi}_{k,r}^{*,s}
\right),
\quad
|\mathcal{S}_{k,r}|\leq 2s .
\end{gathered}
\label{eq:prompt_retrieval_gap}
\end{equation}

\begin{theorem}[Sparse Retrieval Boundary-risk Bound]
\label{thm:propagation_sparse_bound}
Assume $\|\mathbf{p}_j\|_2\leq P_{\max}$ for all prototypes. 
Then
\begin{equation}
\mathcal{R}_{k,r}^{(L)}
\leq
\gamma_{k,r}
\chi_{k,r}^{(L)}
\left[
\epsilon_s(\mathbf{x}_{k,r}^{*})
+
P_{\max}\sqrt{2s}\,
\delta_s(\mathbf{P}_k)
\right].
\label{eq:prism_sparse_risk_bound}
\end{equation}
\end{theorem}

Theorem~\ref{thm:propagation_sparse_bound} shows that boundary risk depends on both bank coverage and retrieval mismatch. 
The coverage term measures whether the global bank contains a sparse semantic basis for the missing modality, while the mismatch term measures whether the topology-conditioned metric selects suitable prototypes. 
Both terms are further amplified by structural exposure, indicating that retrieval errors are most harmful in topology-sensitive regions.

PRISM also budgets the injection strength according to retrieval confidence. 
Let $\Delta S_{k,r}=S_{r,1}-S_{r,2}$ be the retrieval margin between the top-1 and top-2 prototypes, and define
\begin{equation}
\gamma_{k,r}
=
g(\Delta S_{k,r})
=
\operatorname{clip}
\left(
\sigma(a(\Delta S_{k,r}-c_0)),
0,
\gamma_{\max}
\right),
\label{eq:confidence_gate_theory}
\end{equation}
where $g(\cdot)$ is nondecreasing. 
Combining Eq.~\eqref{eq:confidence_gate_theory} with Theorem~\ref{thm:propagation_sparse_bound} yields
\begin{equation}
\mathcal{R}_{k,r}^{(L)}
\leq
g(\Delta S_{k,r})
\chi_{k,r}^{(L)}
\left[
\epsilon_s(\mathbf{x}_{k,r}^{*})
+
P_{\max}\sqrt{2s}\,
\delta_s(\mathbf{P}_k)
\right].
\label{eq:confidence_gate_risk_bound}
\end{equation}
Thus, ambiguous retrievals receive a smaller propagation budget, reducing the chance that unreliable semantics dominate local message passing.

\subsection{Communication-efficient Semantic Exchange}

Finally, we analyze the communication behavior of PRISM. 
This analysis concerns semantic payload rather than communication-frequency reduction. 
PRISM keeps the standard federated optimization schedule, but replaces node-level missing-modality transfer with cluster-level summaries and sparse prototype retrieval.

Let $|\Theta|$ and $|\Phi|$ denote the transmitted shared model and structural prompt parameters. 
For client $k$, let $C_k$ be the number of semantic clusters, $d_q$ the retrieval-key dimension, $d$ the prototype-value dimension, and $s$ the number of retrieved prototypes per cluster. 
Ignoring scalar metadata, the per-round communication of PRISM is
\begin{equation}
\mathcal{C}_{k}^{\mathrm{PRISM}}
=
\mathcal{O}
\left(
|\Theta|+|\Phi|
+
C_k|\mathcal{M}_k|d_q
+
C_k(s+1)d
\right).
\label{eq:prism_total_payload}
\end{equation}

\begin{proposition}[Communication-efficient Semantic Exchange]
\label{prop:communication_payload}
Assume that client $k$ uses $C_k$ semantic clusters and retrieves at most $s$ prototypes per cluster. 
Then the additional semantic communication of PRISM is independent of the number of local nodes $n_k$ and edges $|\mathcal{E}_k|$. 
In contrast, node-level missing-modality feature transfer requires
\begin{equation}
\mathcal{C}_{k}^{\mathrm{node}}
=
\Omega
\left(
n_k|\bar{\mathcal{M}}_k|d
\right).
\label{eq:node_level_comm}
\end{equation}
Therefore, if $C_k(s+1)\ll n_k|\bar{\mathcal{M}}_k|$, PRISM requires strictly smaller semantic communication than node-level missing-modality transfer.
\end{proposition}

Proposition~\ref{prop:communication_payload} shows that PRISM communicates semantic information at the cluster and prototype level. 
Although the server maintains a global bank with $J$ prototypes, each deficient client downloads only the selected prototypes required by sparse retrieval. 
Thus, the client-side semantic communication depends on $C_k$ and $s$, rather than on the local graph size. 
Since PRISM does not skip synchronization rounds, its optimization behavior mainly follows the underlying FGL backbone, while the additional approximation error is governed by the quality of cluster-level retrieval and sparse prototype selection.

Together, the above results provide a compact theoretical justification for PRISM. 
Federation-level retrieval addresses irreducible local semantic uncertainty, topology-aware and confidence-controlled injection limits boundary risk during graph propagation, and cluster-level sparse retrieval reduces semantic communication without requiring node-level missing-modality transfer.

\section{Experiments}
\label{sec:experiments}

We evaluate whether \textbf{PRISM} resolves the central challenge of modality-deficient MM-FGL: deficient clients require cross-client semantic evidence to recover the absent modality basis, while the borrowed semantics must be injected without uncontrolled graph propagation. 
We organize the evaluation around four questions. 
\textbf{EQ1:} Does PRISM consistently outperform strong baselines, especially under client-level modality deficiency? 
\textbf{EQ2:} Which components contribute to the gains, and are topology-aware retrieval and injection necessary? 
\textbf{EQ3:} Is PRISM robust to hyper-parameters, GNN backbones, client scale, and missing-modality severity? 
\textbf{EQ4:} Does PRISM offer favorable convergence and efficiency?

\subsection{Experimental Setup}
\label{sec:experimental_setup}

\textbf{Datasets, tasks, and metrics.}
We evaluate PRISM on six multimodal graph benchmarks: \textit{Toys} and \textit{Grocery} from recommendation graphs~\cite{yan2025graph}, \textit{KU} and \textit{Bili Food} from social/recommendation scenarios~\cite{zhang2024ninerec}, and \textit{QB} and \textit{Cartoon} from e-commerce graphs~\cite{zhang2024ninerec}. 
These datasets contain paired text/image attributes and graph structures. 
We consider node classification, modality matching, and cross-modal retrieval, measured by Accuracy, AUC, and Recall@5, respectively.

\textbf{Federated setting and modality deficiency.}
Each global graph is partitioned into $K$ disjoint client subgraphs using Louvain community detection~\cite{blondel2008fast}, preserving local coherence while inducing label, feature, and topology heterogeneity. 
We report two modality settings: \textit{Full}, where all clients observe all modalities, and \textit{Miss}, where clients are split into full-modal, image-only, and text-only groups with a ratio of $40\%{:}30\%{:}30\%$. 
The latter directly simulates client-level modality deficiency rather than random entry-wise masking.

\textbf{Baselines and implementation.}
We compare PRISM with six groups of baselines: 
(1) General FL: FedAvg~\cite{mcmahan2017communication}; 
(2) MM-GNN: Fed-MGNet~\cite{luanyuan2024mgnet} and Fed-MHGAT~\cite{jia2023multimodal}; 
(3) (MM)-FGL: FedLap~\cite{aliakbari2025subgraph}, S2FGL~\cite{tan2025s2fgl}, and FedProto~\cite{tan2022fedproto}; 
(4) Missing-modality FL: FedMVP~\cite{singha2025fedmvp}, FedMAC~\cite{nguyen2024fedmac}, and MH-pFLID~\cite{xie2024mh}. 
(5) Federated graph augmentation methods: FedSage~\cite{zhang2021subgraph}, FedDEP~\cite{zhang2024deep}, and FedC4~\cite{chen2025rethinking}; 
and (6) Topology-aware FGL methods: FedSPA~\cite{tan2025fedspa} and FedIIH~\cite{yu2025modeling}. 
All methods use the same CLIP-ViT-Large-Patch14 initialization~\cite{radford2021learning,dosovitskiy2020image}, partition protocol, and downstream heads. 
Unless otherwise specified, PRISM adopts a two-layer GCN~\cite{kipf2016semi}, $B=16$ prompt bases, $J=64$ prototypes, top-$s=4$ retrieval, and $T_w=30$ warm-up rounds.

\subsection{Performance Comparison (EQ1)}
\label{sec:overall_performance}

To answer \textbf{EQ1}, we compare PRISM with representative baselines across three downstream tasks under both complete-modality and modality-deficient settings.

\textbf{Overall Performance.} Table~\ref{tab:main_results} reports performance across three tasks and both modality settings. 
PRISM achieves the best result in all columns, with its advantage most evident under \textit{Miss}. 
Across the six missing-modality columns, PRISM improves over the strongest baseline by an average of \textbf{+4.48} points, confirming its effectiveness for modality-deficient clients.
The gains are especially pronounced on modality-centric tasks. 
PRISM improves QB retrieval Recall@5 from \textbf{66.54\%} to \textbf{77.25\%}, and Cartoon retrieval from \textbf{69.37\%} to \textbf{76.25\%}. 
It also yields substantial improvements on KU and Bili Food modality matching. 
These results support the \emph{retrieve globally} component of PRISM: deficient clients benefit from a federation-level semantic basis rather than relying solely on locally incomplete observations.
Moreover, missing-modality FL baselines such as FedMVP and FedMAC remain weaker than PRISM on graph-based tasks, indicating that cross-client semantic transfer alone is insufficient.


\definecolor{orangeL}{rgb}{1.0, 0.95, 0.9}
\definecolor{greenL}{rgb}{0.9, 1.0, 0.9}
\definecolor{blueL}{rgb}{0.9, 0.95, 1.0}
\definecolor{yellowL}{rgb}{1.0, 1.0, 0.9}
\definecolor{grayL}{rgb}{0.95, 0.95, 0.95}
\definecolor{purpleL}{HTML}{E6E6FA}
\definecolor{cyanL}{HTML}{E0FFFF}

\begin{table*}[t]
\centering
\caption{Overall Performance (Mean $\pm$ STD (\%)). \textbf{Bold} indicates the best performance, while \underline{underline} indicates the second best.}
\vspace{-4pt}
\label{tab:main_results}
\resizebox{\textwidth}{!}{
\begin{tabular}{ll cccc cccc cccc}
\toprule
\multirow{3}{*}{\textbf{Category}} & \multirow{3}{*}{\textbf{Method}} & \multicolumn{4}{c}{\textbf{Node Classification (Acc $\uparrow$)}} & \multicolumn{4}{c}{\textbf{Modality Matching (AUC $\uparrow$)}} & \multicolumn{4}{c}{\textbf{Modality Retrieval (R@5 $\uparrow$)}} \\
\cmidrule(lr){3-6} \cmidrule(lr){7-10} \cmidrule(lr){11-14}
& & \multicolumn{2}{c}{\textbf{Toys}} & \multicolumn{2}{c}{\textbf{Grocery}} & \multicolumn{2}{c}{\textbf{KU}} & \multicolumn{2}{c}{\textbf{Bili Food}} & \multicolumn{2}{c}{\textbf{QB}} & \multicolumn{2}{c}{\textbf{Cartoon}} \\
\cmidrule(lr){3-4} \cmidrule(lr){5-6} \cmidrule(lr){7-8} \cmidrule(lr){9-10} \cmidrule(lr){11-12} \cmidrule(lr){13-14}
& & Full & Miss & Full & Miss & Full & Miss & Full & Miss & Full & Miss & Full & Miss \\
\midrule

\cellcolor{orangeL}\textbf{FL} & \cellcolor{orangeL}FedAvg 
& 75.36{\scriptsize$\pm$0.4} & 51.68{\scriptsize$\pm$0.8} & 74.25{\scriptsize$\pm$0.5} & 53.54{\scriptsize$\pm$0.9} 
& 63.52{\scriptsize$\pm$0.6} & 42.26{\scriptsize$\pm$1.1} & 60.85{\scriptsize$\pm$0.7} & 35.25{\scriptsize$\pm$1.3} 
& 73.25{\scriptsize$\pm$0.4} & 56.85{\scriptsize$\pm$0.8} & 71.58{\scriptsize$\pm$0.6} & 54.25{\scriptsize$\pm$0.9} \\
\midrule

\cellcolor{greenL} & \cellcolor{greenL}Fed-MGNet 
& 77.56{\scriptsize$\pm$0.3} & 58.25{\scriptsize$\pm$0.6} & 75.62{\scriptsize$\pm$0.4} & 60.32{\scriptsize$\pm$0.7} 
& 68.62{\scriptsize$\pm$0.5} & 52.68{\scriptsize$\pm$0.9} & 65.23{\scriptsize$\pm$0.6} & 46.85{\scriptsize$\pm$1.0} 
& 78.64{\scriptsize$\pm$0.3} & 62.15{\scriptsize$\pm$0.7} & 76.25{\scriptsize$\pm$0.5} & 60.25{\scriptsize$\pm$0.8} \\
\multirow{-2}{*}{\cellcolor{greenL}\textbf{MM-GNN}} & \cellcolor{greenL}Fed-MHGAT 
& \underline{80.25}{\scriptsize$\pm$0.2} & 59.15{\scriptsize$\pm$0.5} & \underline{79.62}{\scriptsize$\pm$0.3} & 61.85{\scriptsize$\pm$0.7} 
& 69.51{\scriptsize$\pm$0.4} & 53.26{\scriptsize$\pm$0.8} & 68.25{\scriptsize$\pm$0.5} & 48.57{\scriptsize$\pm$0.9} 
& 81.75{\scriptsize$\pm$0.3} & 65.84{\scriptsize$\pm$0.6} & 79.85{\scriptsize$\pm$0.4} & 63.85{\scriptsize$\pm$0.7} \\
\midrule

\cellcolor{blueL} & \cellcolor{blueL}FedLap 
& 77.56{\scriptsize$\pm$0.4} & 54.55{\scriptsize$\pm$0.7} & 76.55{\scriptsize$\pm$0.5} & 55.48{\scriptsize$\pm$0.8} 
& 65.88{\scriptsize$\pm$0.6} & 56.51{\scriptsize$\pm$0.5} & 62.85{\scriptsize$\pm$0.7} & 54.17{\scriptsize$\pm$0.6} 
& 74.65{\scriptsize$\pm$0.5} & 54.55{\scriptsize$\pm$0.9} & 72.95{\scriptsize$\pm$0.6} & 52.97{\scriptsize$\pm$1.0} \\
\cellcolor{blueL} & \cellcolor{blueL}S2FGL 
& 79.58{\scriptsize$\pm$0.3} & 56.28{\scriptsize$\pm$0.6} & 78.56{\scriptsize$\pm$0.4} & 58.58{\scriptsize$\pm$0.7} 
& 66.87{\scriptsize$\pm$0.9} & 46.85{\scriptsize$\pm$0.9} & 63.75{\scriptsize$\pm$1.2} & 44.67{\scriptsize$\pm$1.2} 
& 76.87{\scriptsize$\pm$0.4} & 56.85{\scriptsize$\pm$0.8} & 74.96{\scriptsize$\pm$0.5} & 54.37{\scriptsize$\pm$0.9} \\
\multirow{-3}{*}{\cellcolor{blueL}\textbf{(MM)-FGL}} & \cellcolor{blueL}FedProto 
& 79.62{\scriptsize$\pm$0.2} & 65.24{\scriptsize$\pm$0.5} & 79.23{\scriptsize$\pm$0.3} & 68.32{\scriptsize$\pm$0.6} 
& 71.15{\scriptsize$\pm$0.4} & 58.52{\scriptsize$\pm$0.8} & 68.15{\scriptsize$\pm$0.5} & 53.54{\scriptsize$\pm$0.9} 
& 79.64{\scriptsize$\pm$0.4} & 62.68{\scriptsize$\pm$0.3} & 71.89{\scriptsize$\pm$0.8} & 69.37{\scriptsize$\pm$0.5} \\
\midrule

\cellcolor{yellowL} & \cellcolor{yellowL}FedMVP 
& 78.95{\scriptsize$\pm$0.4} & 67.65{\scriptsize$\pm$0.3} & 74.85{\scriptsize$\pm$0.8} & 65.25{\scriptsize$\pm$0.7} 
& 66.81{\scriptsize$\pm$0.7} & 56.51{\scriptsize$\pm$0.7} & 64.37{\scriptsize$\pm$0.8} & 44.58{\scriptsize$\pm$0.8} 
& 81.97{\scriptsize$\pm$0.3} & 66.54{\scriptsize$\pm$0.6} & 79.68{\scriptsize$\pm$0.4} & 64.35{\scriptsize$\pm$0.8} \\
\cellcolor{yellowL} & \cellcolor{yellowL}FedMAC 
& 77.85{\scriptsize$\pm$0.4} & 55.98{\scriptsize$\pm$0.7} & 76.58{\scriptsize$\pm$0.5} & 57.85{\scriptsize$\pm$0.8} 
& 70.34{\scriptsize$\pm$0.5} & 47.68{\scriptsize$\pm$1.0} & 68.35{\scriptsize$\pm$0.6} & 44.68{\scriptsize$\pm$1.1} 
& 80.95{\scriptsize$\pm$0.4} & 63.59{\scriptsize$\pm$0.7} & 78.68{\scriptsize$\pm$0.5} & 61.97{\scriptsize$\pm$0.9} \\
\multirow{-3}{*}{\cellcolor{yellowL}\textbf{Miss-Modality}} & \cellcolor{yellowL}MH-pFLID 
& 78.52{\scriptsize$\pm$0.4} & 54.85{\scriptsize$\pm$0.8} & 77.52{\scriptsize$\pm$0.5} & 56.25{\scriptsize$\pm$0.8} 
& 71.35{\scriptsize$\pm$0.4} & 53.24{\scriptsize$\pm$0.9} & 69.35{\scriptsize$\pm$0.5} & 50.17{\scriptsize$\pm$1.1} 
& 79.65{\scriptsize$\pm$0.4} & 65.12{\scriptsize$\pm$0.7} & 77.24{\scriptsize$\pm$0.6} & 63.32{\scriptsize$\pm$0.8} \\
\midrule

\cellcolor{purpleL} & \cellcolor{purpleL}FedSage 
& 76.12{\scriptsize$\pm$0.5} & 57.85{\scriptsize$\pm$0.8} & 75.30{\scriptsize$\pm$0.6} & 59.45{\scriptsize$\pm$0.7} 
& 67.42{\scriptsize$\pm$0.6} & 50.15{\scriptsize$\pm$1.0} & 64.12{\scriptsize$\pm$0.7} & 45.32{\scriptsize$\pm$1.1} 
& 77.50{\scriptsize$\pm$0.5} & 59.65{\scriptsize$\pm$0.8} & 75.40{\scriptsize$\pm$0.6} & 60.15{\scriptsize$\pm$0.9} \\
\cellcolor{purpleL} & \cellcolor{purpleL}FedDEP 
& 77.85{\scriptsize$\pm$0.4} & 61.25{\scriptsize$\pm$0.6} & 76.85{\scriptsize$\pm$0.5} & 62.15{\scriptsize$\pm$0.7} 
& 68.95{\scriptsize$\pm$0.5} & 54.65{\scriptsize$\pm$0.9} & 66.50{\scriptsize$\pm$0.6} & 49.85{\scriptsize$\pm$1.0} 
& 78.65{\scriptsize$\pm$0.4} & 62.45{\scriptsize$\pm$0.7} & 76.85{\scriptsize$\pm$0.5} & 64.30{\scriptsize$\pm$0.8} \\
\multirow{-3}{*}{\cellcolor{purpleL}\textbf{Graph Aug.}} & \cellcolor{purpleL}FedC4 
& 79.15{\scriptsize$\pm$0.3} & 68.45{\scriptsize$\pm$0.5} & 78.45{\scriptsize$\pm$0.4} & 69.15{\scriptsize$\pm$0.6} 
& 70.85{\scriptsize$\pm$0.4} & 60.25{\scriptsize$\pm$0.7} & 68.75{\scriptsize$\pm$0.5} & 56.85{\scriptsize$\pm$0.8} 
& 80.50{\scriptsize$\pm$0.4} & 68.50{\scriptsize$\pm$0.6} & 78.95{\scriptsize$\pm$0.5} & 71.45{\scriptsize$\pm$0.7} \\
\midrule

\cellcolor{cyanL} & \cellcolor{cyanL}FedSPA 
& 79.82{\scriptsize$\pm$0.3} & 67.92{\scriptsize$\pm$0.5} & 78.95{\scriptsize$\pm$0.4} & 68.55{\scriptsize$\pm$0.6} 
& 72.15{\scriptsize$\pm$0.4} & 59.35{\scriptsize$\pm$0.7} & 68.92{\scriptsize$\pm$0.6} & 55.42{\scriptsize$\pm$0.8} 
& 81.50{\scriptsize$\pm$0.4} & 67.25{\scriptsize$\pm$0.6} & 79.12{\scriptsize$\pm$0.5} & 70.25{\scriptsize$\pm$0.7} \\
\multirow{-2}{*}{\cellcolor{cyanL}\textbf{Topo-FGL}} & \cellcolor{cyanL}FedIIH 
& 80.12{\scriptsize$\pm$0.2} & \underline{68.95}{\scriptsize$\pm$0.4} & 79.35{\scriptsize$\pm$0.3} & \underline{69.85}{\scriptsize$\pm$0.5} 
& \underline{72.85}{\scriptsize$\pm$0.4} & \underline{61.42}{\scriptsize$\pm$0.6} & \underline{70.15}{\scriptsize$\pm$0.5} & \underline{58.25}{\scriptsize$\pm$0.7} 
& \underline{82.45}{\scriptsize$\pm$0.3} & \underline{70.15}{\scriptsize$\pm$0.5} & \underline{81.25}{\scriptsize$\pm$0.4} & \underline{72.85}{\scriptsize$\pm$0.6} \\
\midrule

\rowcolor{grayL}\textbf{Ours} & \textbf{PRISM} 
& \textbf{80.65}{\scriptsize$\pm$0.2} & \textbf{70.28}{\scriptsize$\pm$0.4} & \textbf{79.64}{\scriptsize$\pm$0.3} & \textbf{71.58}{\scriptsize$\pm$0.5} 
& \textbf{74.94}{\scriptsize$\pm$0.3} & \textbf{66.51}{\scriptsize$\pm$0.5} & \textbf{72.59}{\scriptsize$\pm$0.4} & \textbf{63.14}{\scriptsize$\pm$0.6} 
& \textbf{87.26}{\scriptsize$\pm$0.2} & \textbf{77.25}{\scriptsize$\pm$0.4} & \textbf{85.46}{\scriptsize$\pm$0.3} & \textbf{76.25}{\scriptsize$\pm$0.5} \\
\bottomrule
\end{tabular}
}
\end{table*}

\subsection{Ablation Study (EQ2)}
\label{sec:ablation}

\begin{table}[!t]
    \centering
    \caption{Ablation study of \textbf{PRISM} on three datasets. We report the mean and standard deviation over 5 independent runs.}
    \vspace{-4pt}
    \label{table:ablation_study}
    \resizebox{\columnwidth}{!}{
    \begin{tabular}{l|cc|cc|cc}
    \toprule
    \multirow{2}{*}{\textbf{Variant}} & \multicolumn{2}{c|}{\textbf{Dataset: Toys (Acc $\uparrow$)}} & \multicolumn{2}{c|}{\textbf{Dataset: KU (AUC $\uparrow$)}} & \multicolumn{2}{c}{\textbf{Dataset: QB (Recall@5 $\uparrow$)}} \\
    \cmidrule(lr){2-3} \cmidrule(lr){4-5} \cmidrule(lr){6-7}
    & \textbf{Full} & \textbf{Miss} & \textbf{Full} & \textbf{Miss} & \textbf{Full} & \textbf{Miss} \\
    \midrule
    \rowcolor[HTML]{F2F2F2} 
    \textbf{Full PRISM (Ours)} & \textbf{80.65$\pm$0.2} & \textbf{70.28$\pm$0.4} & \textbf{74.94$\pm$0.3} & \textbf{66.51$\pm$0.5} & \textbf{87.26$\pm$0.3} & \textbf{77.25$\pm$0.4} \\
    \midrule
    w/o Low-Rank Prompt Basis                   & 76.82$\pm$0.5 & 65.57$\pm$0.8 & 69.85$\pm$0.6 & 63.40$\pm$0.9 & 82.35$\pm$0.5 & 72.50$\pm$0.7 \\ 
    w/o Topology-Conditioned Projection ($P_k$) & 77.40$\pm$0.4 & 64.51$\pm$0.7 & 70.30$\pm$0.5 & 62.15$\pm$0.8 & 83.10$\pm$0.4 & 71.20$\pm$0.6 \\
    w/o Confidence-Aware Gating ($\gamma_r$)    & 78.95$\pm$0.3 & 62.90$\pm$0.9 & 71.50$\pm$0.4 & 60.20$\pm$0.9 & 84.20$\pm$0.3 & 69.80$\pm$0.8 \\
    w/o Proactive Retrieval                     & 74.36$\pm$0.6 & 55.24$\pm$1.1 & 66.40$\pm$0.7 & 55.30$\pm$1.2 & 80.50$\pm$0.6 & 62.40$\pm$0.9 \\
    w/ Shuffled Structural Prior ($z_k$)        & 78.12$\pm$0.4 & 63.86$\pm$0.8 & 71.08$\pm$0.5 & 61.42$\pm$0.8 & 83.74$\pm$0.4 & 70.63$\pm$0.7 \\
    w/ Topology-Agnostic Injection              & 77.85$\pm$0.5 & 63.18$\pm$0.9 & 70.76$\pm$0.6 & 60.95$\pm$0.9 & 83.26$\pm$0.5 & 70.11$\pm$0.8 \\
    Retrieval + Feature Concatenation           & 77.22$\pm$0.5 & 62.74$\pm$0.9 & 70.21$\pm$0.6 & 60.48$\pm$1.0 & 82.91$\pm$0.5 & 69.56$\pm$0.8 \\
    \bottomrule
    \end{tabular}}
\end{table}

To answer \textbf{EQ2}, we ablate the key components of PRISM and further compare it with diagnostic variants that replace topology-aware retrieval or structural injection with simpler alternatives.

\textbf{Module Ablation.} Table~\ref{table:ablation_study} evaluates the contribution of each component. 
Removing proactive retrieval causes the largest degradation, especially on missing-modality clients, confirming that an absent modality basis cannot be recovered through local graph learning alone. 
Removing the low-rank prompt basis or topology-conditioned projection $P_k$ also consistently weakens performance, showing that retrieval must adapt to the receiving graph rather than rely on a fixed global metric. 
Disabling confidence-aware gating further harms robustness, validating the need to regulate ambiguous semantics before message passing.

\textbf{Comparison with Injection Variants.}
The diagnostic alternatives lead to the same conclusion. Shuffled structural prompts, topology-agnostic anchors, and direct retrieval-plus-feature-concatenation all underperform the full model.

\subsection{Robustness Analysis (EQ3)}
\label{sec:robustness_analysis}

\begin{figure}[!t]
    \centering

    \begin{minipage}[b]{0.45\linewidth}
        \centering
        {\scriptsize (a) Sensitivity of warmup steps ($T_w$)} \\[1ex] 
        \includegraphics[width=1.0\linewidth]{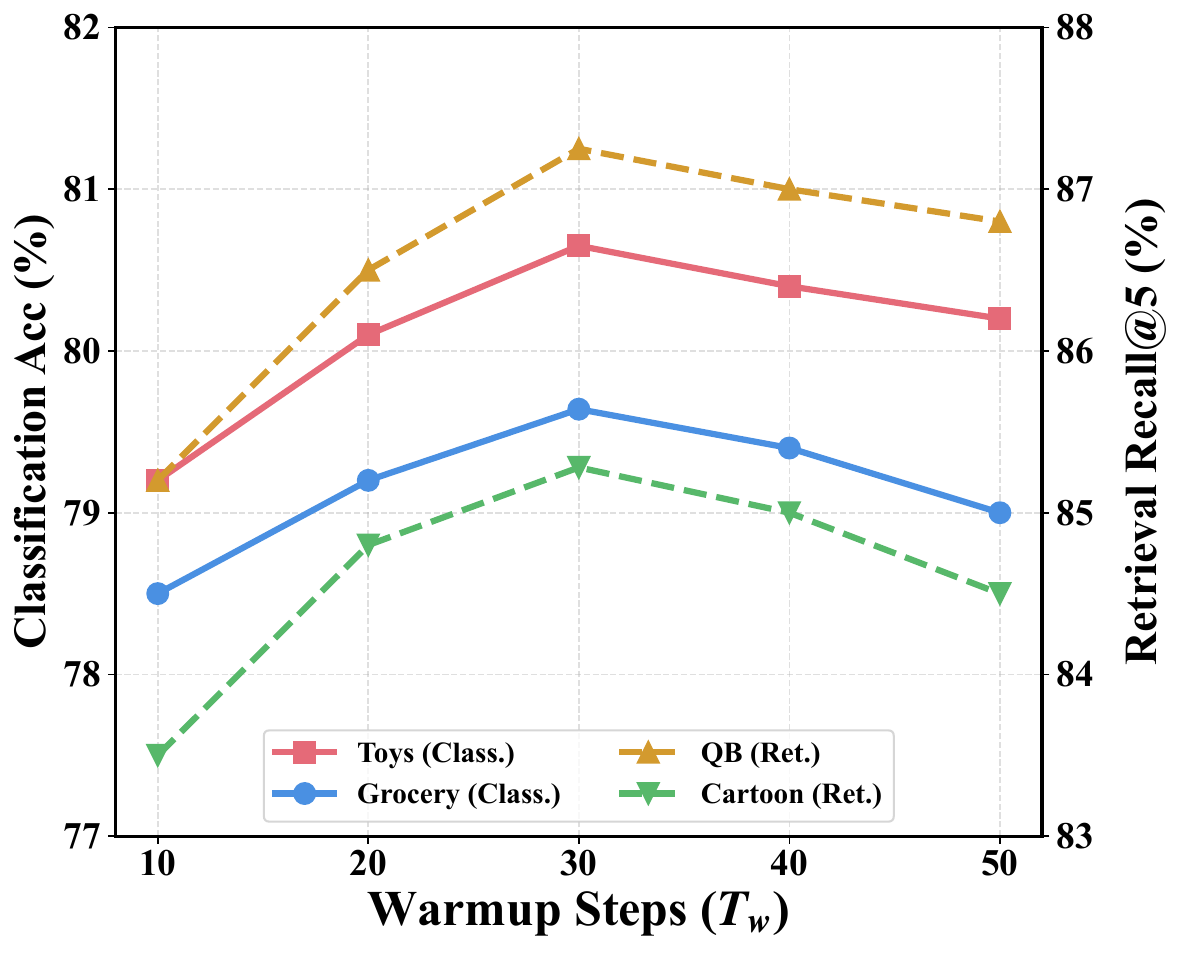}
        \label{fig:sens_w}
    \end{minipage}
    \hfill
    \begin{minipage}[b]{0.45\linewidth}
        \centering
        {\scriptsize (b) Interaction of $B$ and $J$ on Grocery} \\[1ex] 
        \includegraphics[width=1.0\linewidth]{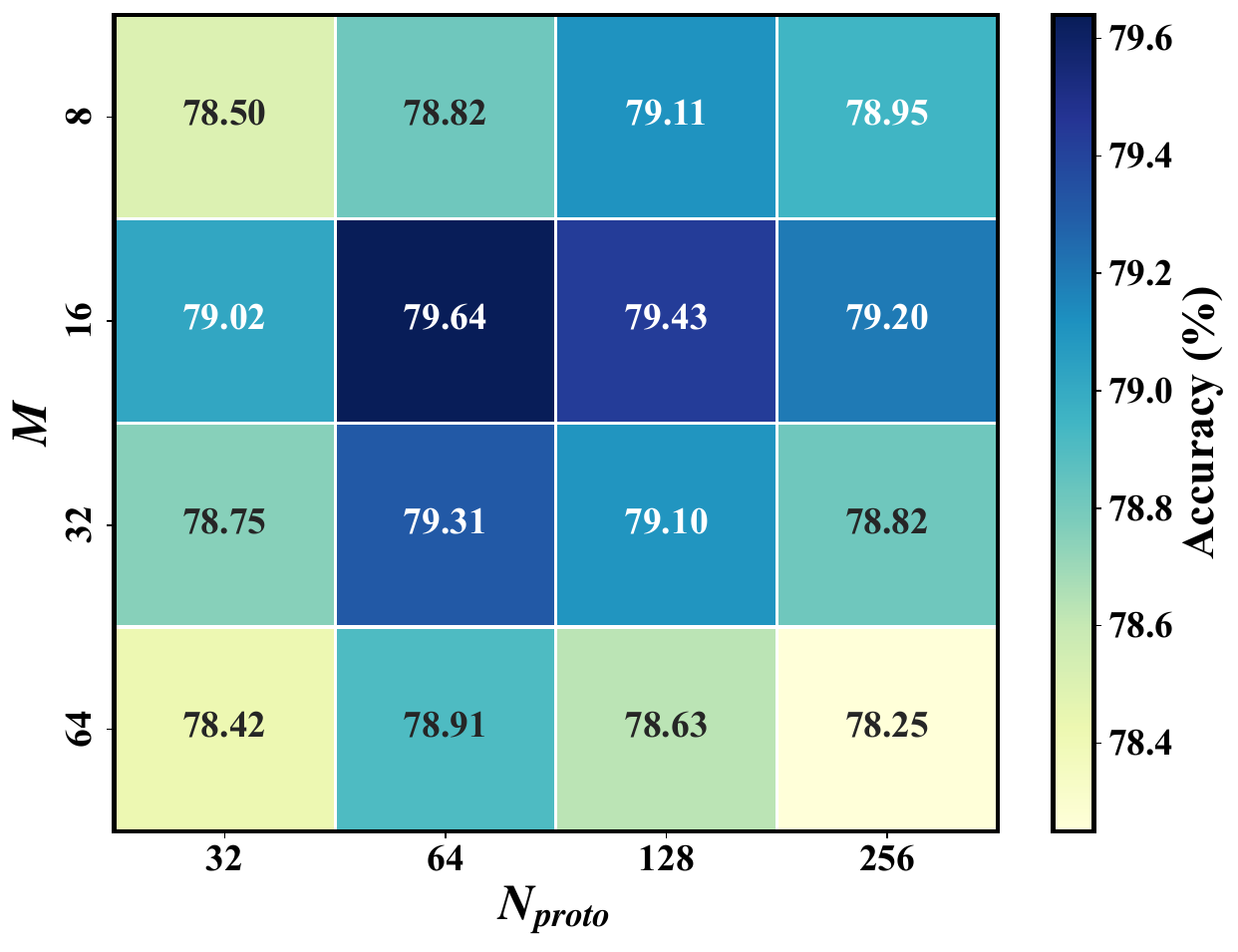}
        \label{fig:sens_heatmap}
    \end{minipage}
    \vspace{-20pt}
    \caption{\textbf{Hyper-parameter sensitivity analysis of PRISM.} In (a), the left y-axis denotes Node Classification Accuracy (\%), and the right y-axis represents Modal Retrieval Recall@5 (\%). (b) displays the interaction between the number of prompt basis elements $B$ and prototypes $J$.}
    \label{fig:sensitivity_all}
\end{figure}

\begin{table}[!t] 
\centering
\caption{\textbf{Performance Comparison across Different GNN Backbones.} We report the Node Classification Accuracy (\%) on the Toys and Grocery datasets under the \textit{Missing} setting.}
\vspace{-4pt}
\label{tab:backbone_ablation}
\resizebox{\columnwidth}{!}{
\begin{tabular}{ll ccccc c} 
\toprule
\textbf{Backbone} & \textbf{Dataset} & \textbf{FedAvg} & \textbf{Fed-MGNet} & \textbf{FedProto} & \textbf{FedMVP} & \textbf{Ours} & \textbf{Gain ($\Delta$)} \\
\midrule

\multirow{2}{*}{\textbf{GCN}}  
& Toys & 51.68{\scriptsize$\pm$0.8} & 58.25{\scriptsize$\pm$0.6} & 65.24{\scriptsize$\pm$0.5} & \underline{67.65}{\scriptsize$\pm$0.3} & \textbf{70.28}{\scriptsize$\pm$0.4} & \cellcolor{green!10}+2.63 \\
& Grocery & 53.54{\scriptsize$\pm$0.9} &60.32{\scriptsize$\pm$0.7} & \underline{68.32}{\scriptsize$\pm$0.6} & 65.25{\scriptsize$\pm$0.7} & \textbf{71.58}{\scriptsize$\pm$0.5} & \cellcolor{green!10}+2.16 \\
\midrule

\multirow{2}{*}{\textbf{GAT}}  
& Toys & 52.45{\scriptsize$\pm$0.7} & 59.18{\scriptsize$\pm$0.5} & 66.15{\scriptsize$\pm$0.4} & \underline{68.52}{\scriptsize$\pm$0.3} & \textbf{71.12}{\scriptsize$\pm$0.3} & \cellcolor{green!10}+2.60 \\
& Grocery & 54.12{\scriptsize$\pm$0.8} & 61.35{\scriptsize$\pm$0.6} & 69.05{\scriptsize$\pm$0.5} & \underline{70.18}{\scriptsize$\pm$0.4} & \textbf{72.35}{\scriptsize$\pm$0.4} & \cellcolor{green!10}+2.17 \\
\midrule

\multirow{2}{*}{\textbf{SAGE}}  
& Toys & 50.92{\scriptsize$\pm$0.9} & 57.85{\scriptsize$\pm$0.8} & 64.88{\scriptsize$\pm$0.6} & \underline{67.12}{\scriptsize$\pm$0.5} & \textbf{69.85}{\scriptsize$\pm$0.5} & \cellcolor{green!10}+2.73 \\
& Grocery & 52.88{\scriptsize$\pm$1.0} & 59.94{\scriptsize$\pm$0.9} & 67.50{\scriptsize$\pm$0.7} & \underline{68.85}{\scriptsize$\pm$0.6} & \textbf{70.94}{\scriptsize$\pm$0.6} & \cellcolor{green!10}+2.09 \\
\midrule

\multirow{2}{*}{\textbf{GIN}}  
& Toys & 49.75{\scriptsize$\pm$1.1} & 56.32{\scriptsize$\pm$0.9} & 63.20{\scriptsize$\pm$0.8} & \underline{65.85}{\scriptsize$\pm$0.7} & \textbf{68.42}{\scriptsize$\pm$0.7} & \cellcolor{green!10}+2.57 \\
& Grocery & 51.36{\scriptsize$\pm$1.2} & 58.75{\scriptsize$\pm$1.0} & 66.12{\scriptsize$\pm$0.9} & \underline{67.28}{\scriptsize$\pm$0.8} & \textbf{69.55}{\scriptsize$\pm$0.8} & \cellcolor{green!10}+2.27 \\

\bottomrule
\end{tabular}
}
\end{table}

\begin{figure}[!t]
    \centering
    \begin{minipage}[b]{0.3\linewidth}
        \centering
        {\tiny \hspace*{3pt} (a) Node Classification (Toys)} \\[1ex]
        \includegraphics[width=\linewidth]{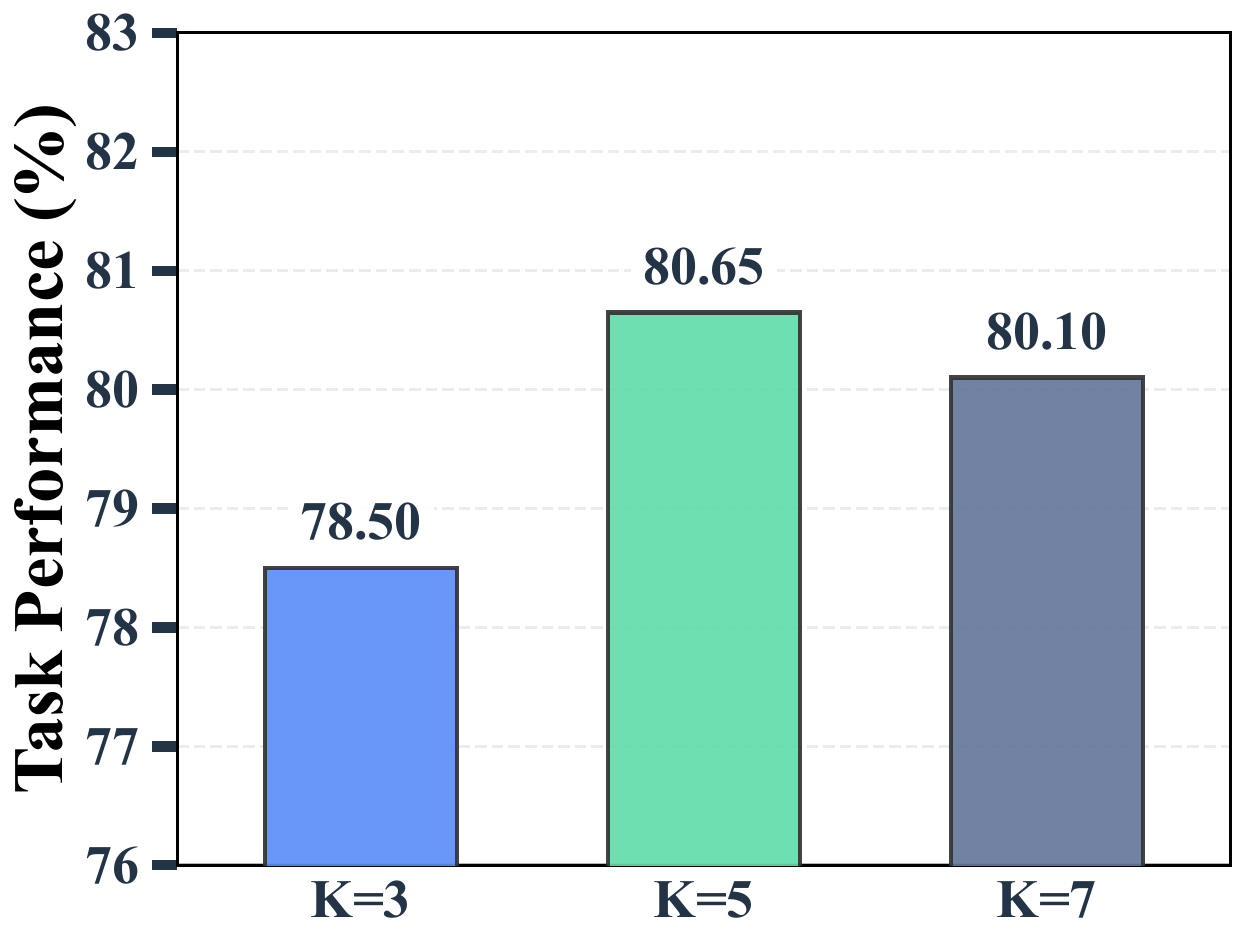}\\[4pt]
        \includegraphics[width=\linewidth]{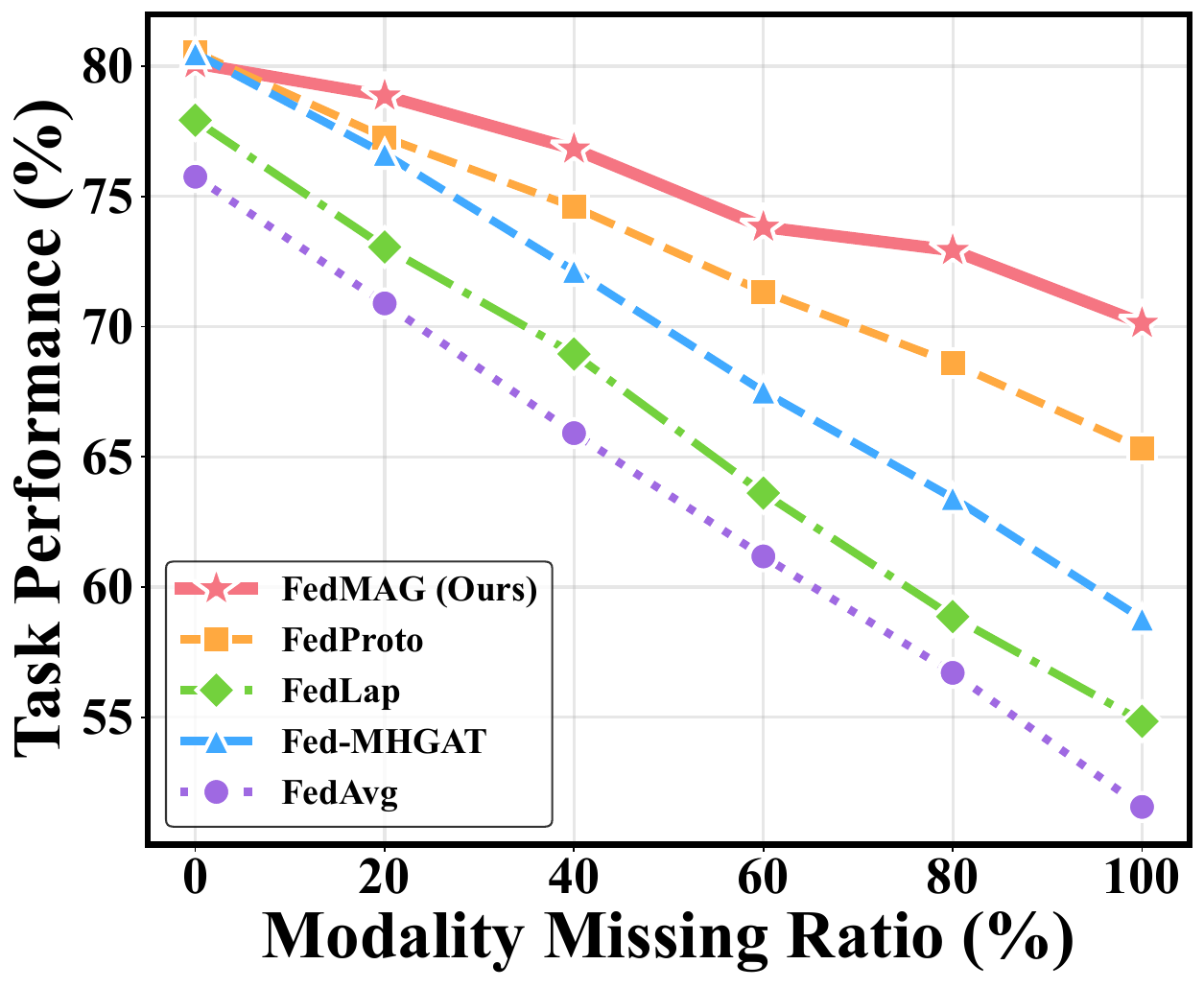}
        \label{fig:col_toys}
    \end{minipage}
    \hfill
    \begin{minipage}[b]{0.3\linewidth}
        \centering
        {\tiny \hspace*{3pt} (b) Modality Matching (KU)} \\[1ex]
        \includegraphics[width=\linewidth]{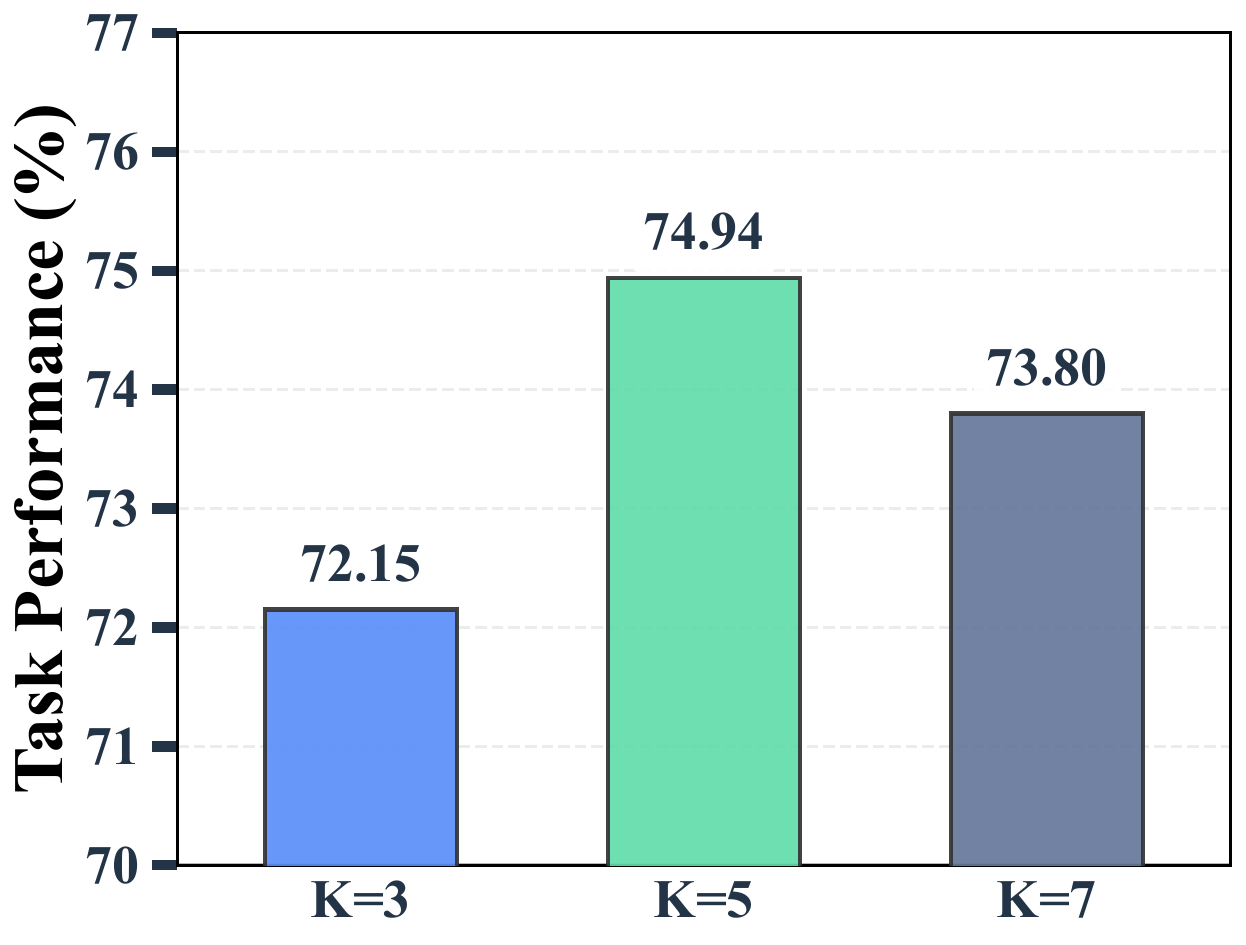}\\[4pt]
        \includegraphics[width=\linewidth]{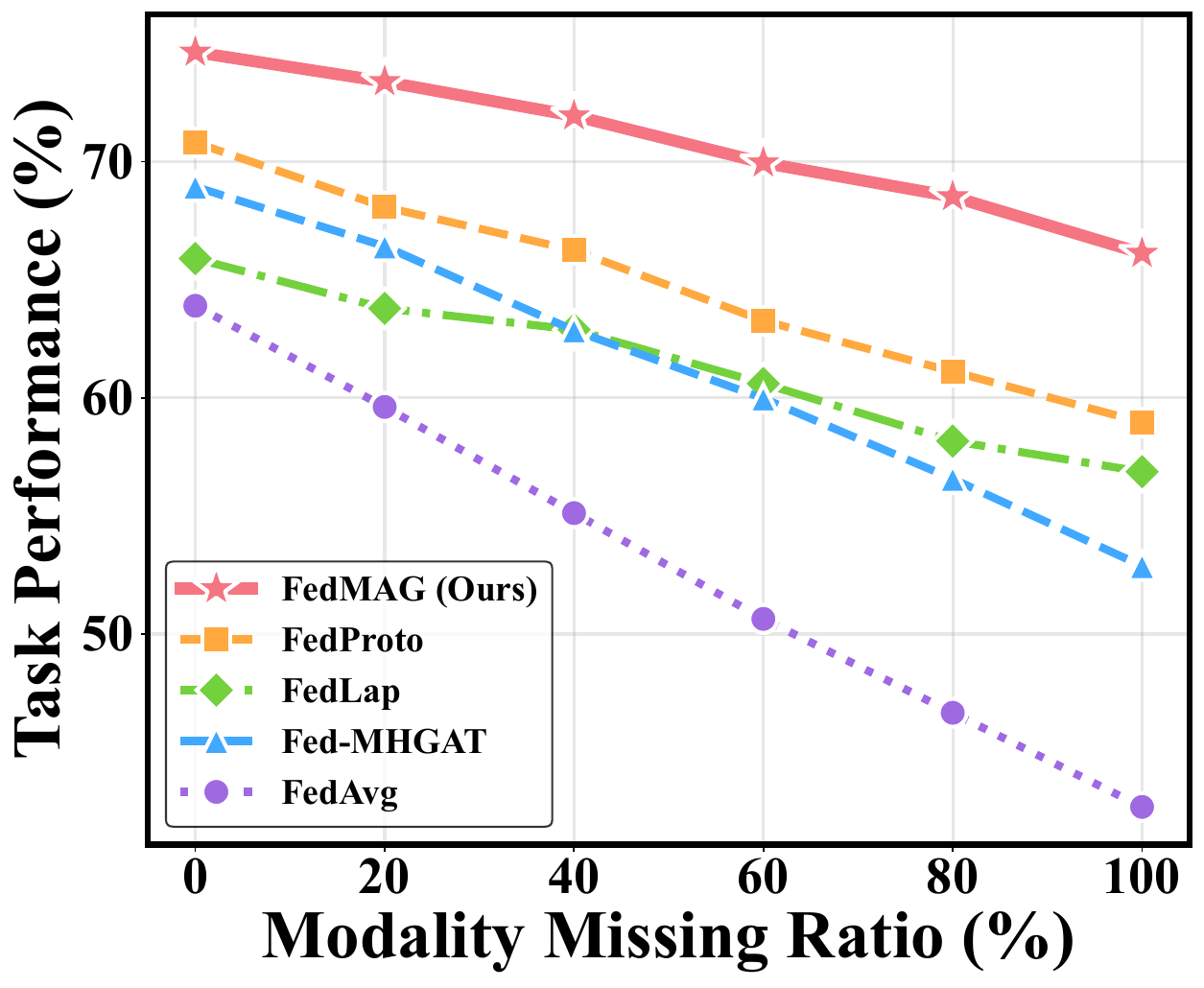}
        \label{fig:col_ku}
    \end{minipage}
    \hfill
    \begin{minipage}[b]{0.3\linewidth}
        \centering
        {\tiny \hspace*{3pt} (c) Modal Retrieval (QB)} \\[1ex]
        \includegraphics[width=\linewidth]{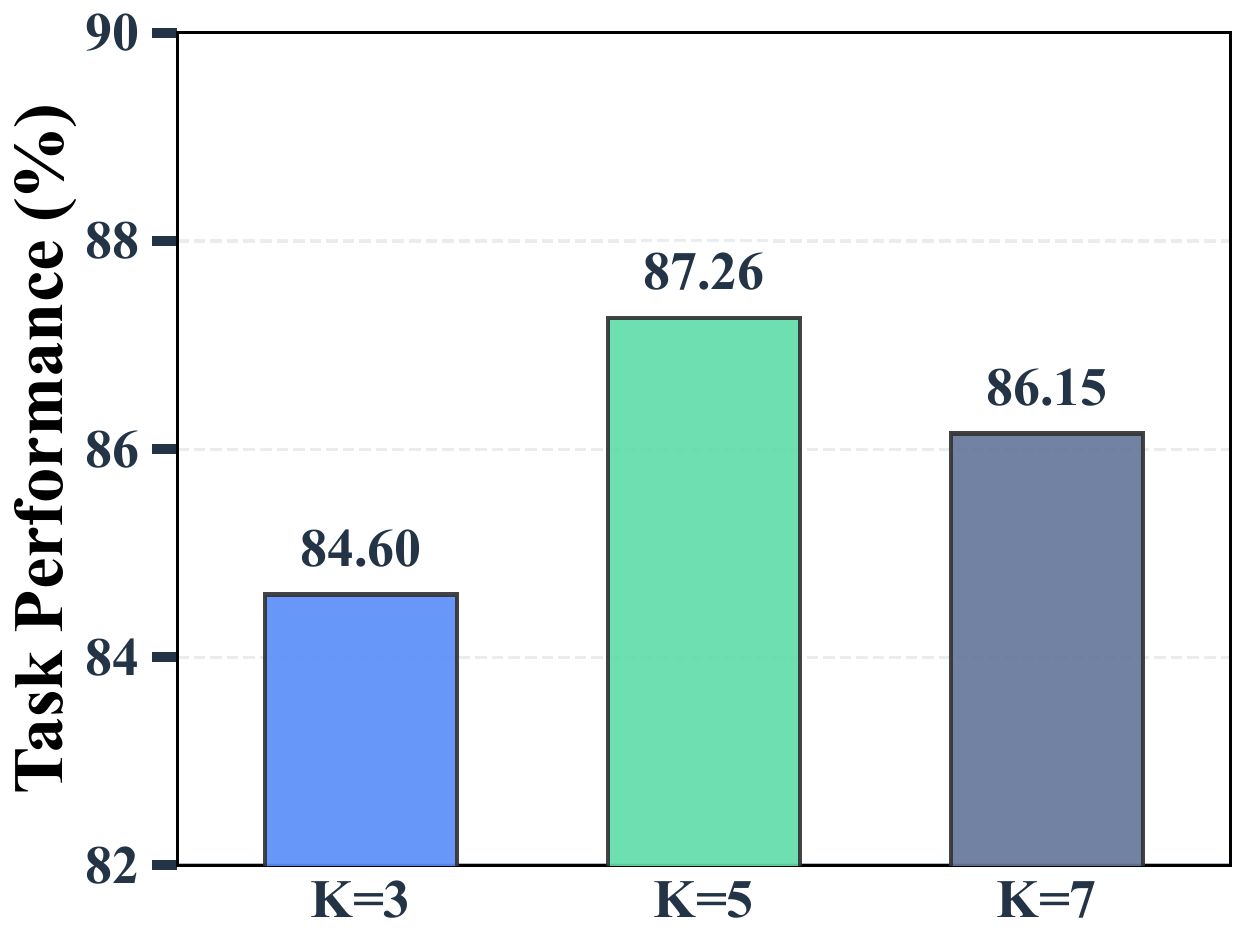}\\[4pt]
        \includegraphics[width=\linewidth]{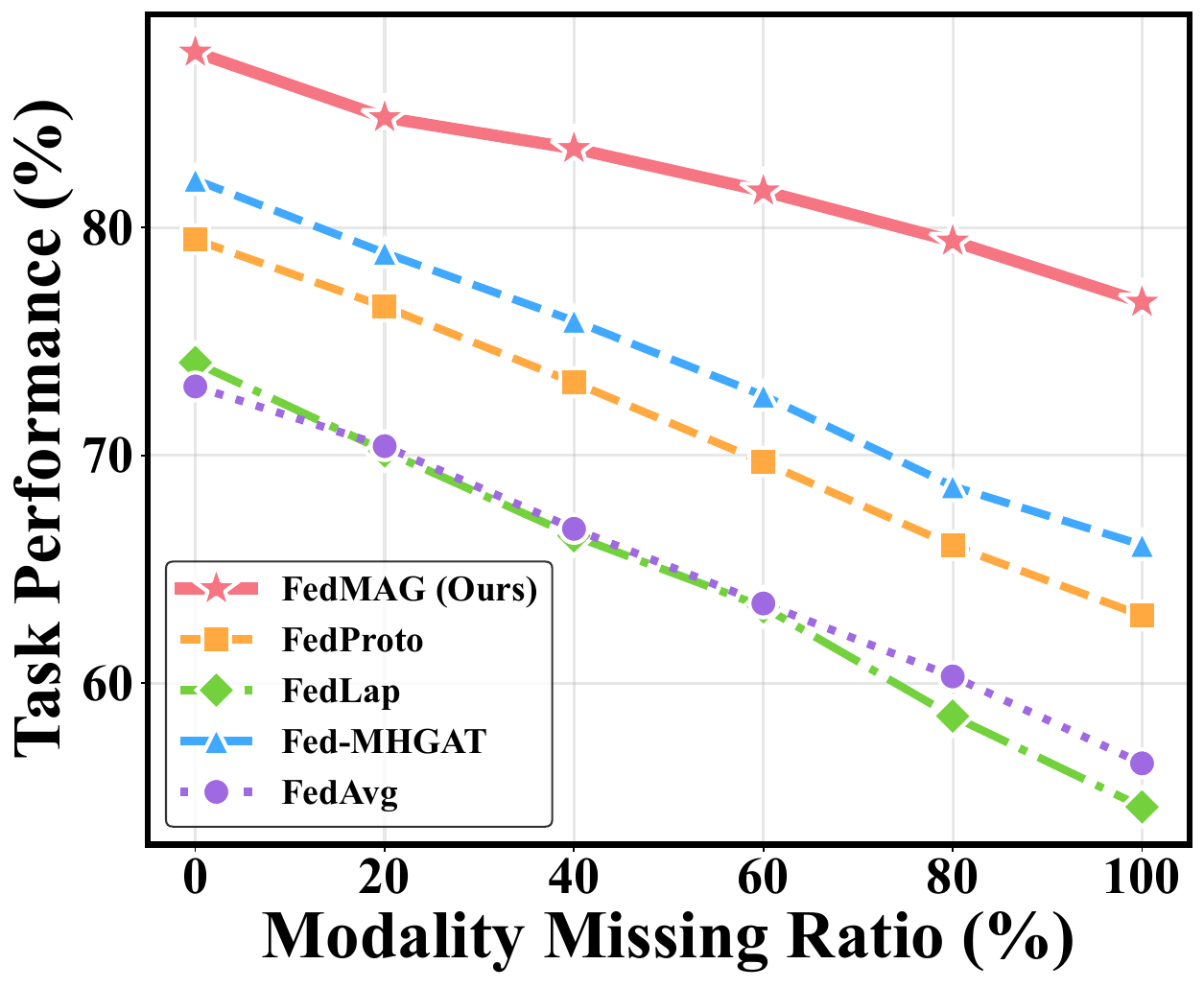}
        \label{fig:col_qb}
    \end{minipage}
    \vspace{-20pt}
    \caption{\textbf{Resilience to Modality Incompleteness.} The top row shows performance as the number of clients $K$ increases from 3 to 7. The bottom row shows performance under modality missing rates from $0\%$ to $100\%$.}
    \label{fig:robustness_results}
\end{figure}

To answer \textbf{EQ3}, we evaluate PRISM under varying hyper-parameters, GNN backbones, client scales, modality-deficiency severities, and asynchronous update scenarios.

\begin{figure}[!t]
    \centering
    \includegraphics[width=0.9\linewidth]{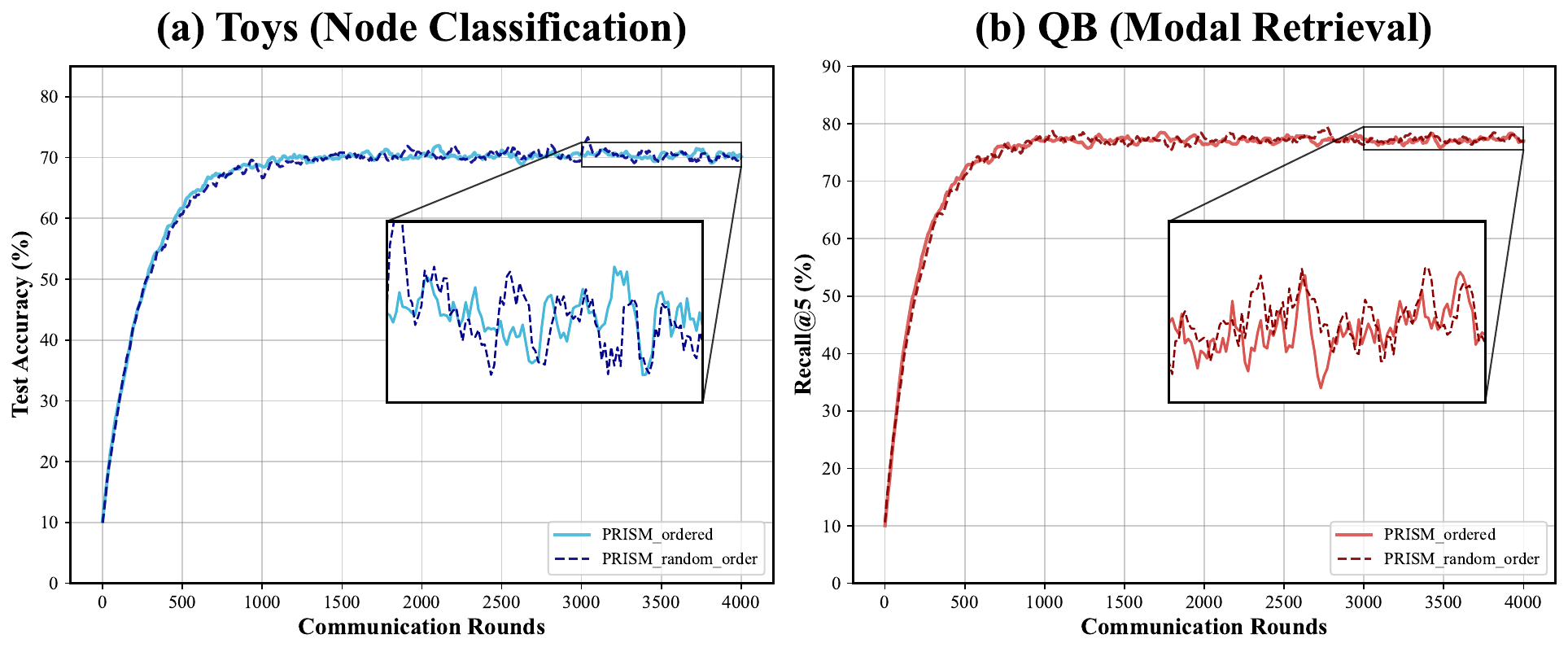}
    \vspace{-10pt}
    \caption{\textbf{Resilience to Asynchronous Client Updates.} Comparison of model accuracy under ordered versus randomly ordered client updates on (a) Toys (Node Classification) and (b) QB (Modal Retrieval) under the \textit{Missing} setting.}
    \label{fig:async_robustness}
\end{figure}

\textbf{Hyperparameter Sensitivity.} Fig.~\ref{fig:sensitivity_all} shows that PRISM remains stable across a broad range of warm-up lengths, prompt-basis sizes, and prototype-bank sizes. 
The setting $T_w=30$, $B=16$, and $J=64$ provides a favorable balance between semantic coverage and retrieval ambiguity.

\textbf{GNN Backbones.}
Table~\ref{tab:backbone_ablation} further evaluates PRISM with GCN~\cite{kipf2016semi}, GAT~\cite{velivckovic2017graph}, GraphSAGE~\cite{hamilton2017inductive}, and GIN~\cite{xu2018powerful} backbones. 
PRISM consistently achieves the best performance across all settings, with gains of \textbf{+2.09} to \textbf{+2.73} points over the strongest baseline. 
This confirms that the improvement stems from the retrieval-and-injection mechanism rather than a specific graph encoder.

\textbf{Fragmentation and Deficiency.}
Finally, Fig.~\ref{fig:robustness_results} examines client scale and missing-modality severity. 
PRISM maintains a clear lead as the number of clients increases from $K=3$ to $K=7$, and it degrades more slowly than competing methods as the missing-modality ratio rises from $0\%$ to $100\%$. 
These results demonstrate strong robustness under both federation fragmentation and severe modality deficiency.

\textbf{Resilience to Asynchronous Client Updates.} 
To validate the effectiveness of PRISM in practical asynchronous federated environments, we compare model performance under ordered versus randomly ordered client updates. As shown in Fig.~\ref{fig:async_robustness}, the resulting convergence trajectories and final task performances are nearly identical on both the Toys and QB datasets under the \textit{Missing} setting. This confirms that PRISM can reliably aggregate cross-client multimodal evidence asynchronously, mitigating the straggler effect and maximizing training efficiency without sacrificing semantic recovery accuracy.

\subsection{Efficiency and Convergence Analysis (EQ4)}
\label{sec:efficiency_analysis}

\begin{figure}[!t]
    \centering
    
    \begin{minipage}[b]{0.3\linewidth}
        \centering
        {\tiny \hspace*{3pt} (a) Node Classification (Toys)} \\[1ex]
        \includegraphics[width=\linewidth]{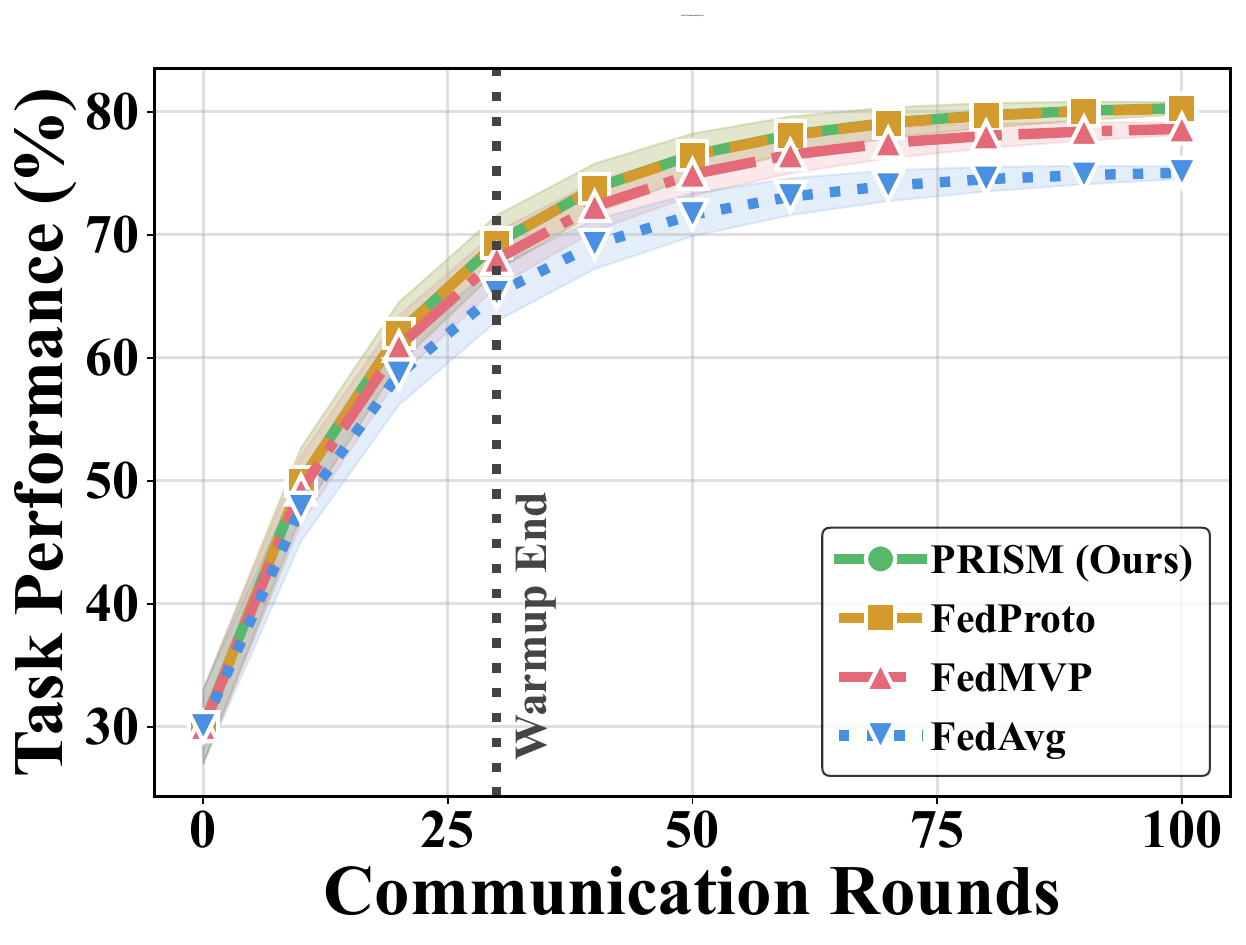}
        \label{fig:conv_toys}
    \end{minipage}
    \hfill
    \begin{minipage}[b]{0.3\linewidth}
        \centering
        {\tiny \hspace*{3pt} (b) Modality Matching (KU)} \\[1ex]
        \includegraphics[width=\linewidth]{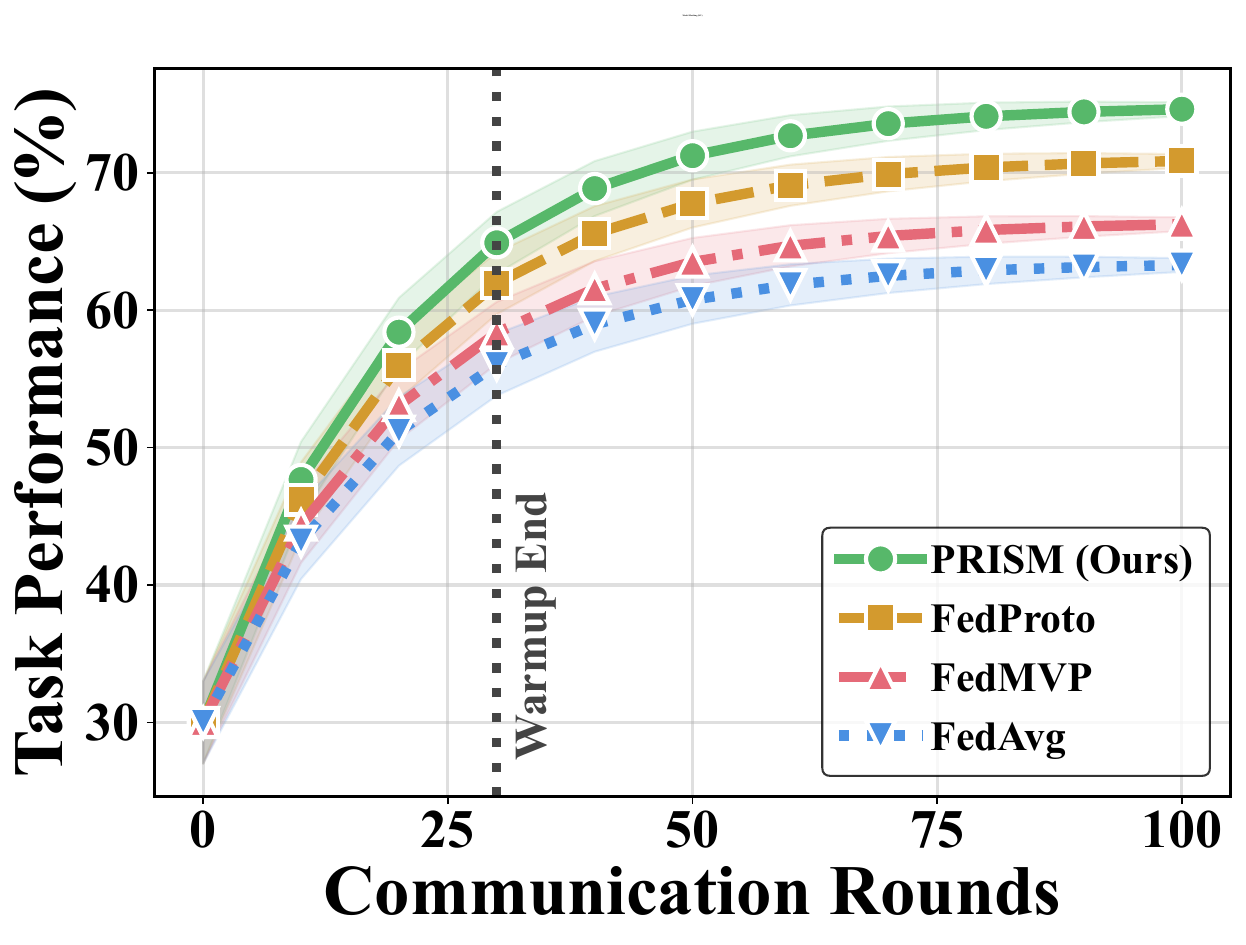}
        \label{fig:conv_ku}
    \end{minipage}
    \hfill
    \begin{minipage}[b]{0.3\linewidth}
        \centering
        {\tiny \hspace*{3pt} (c) Modal Retrieval (QB)} \\[1ex]
        \includegraphics[width=\linewidth]{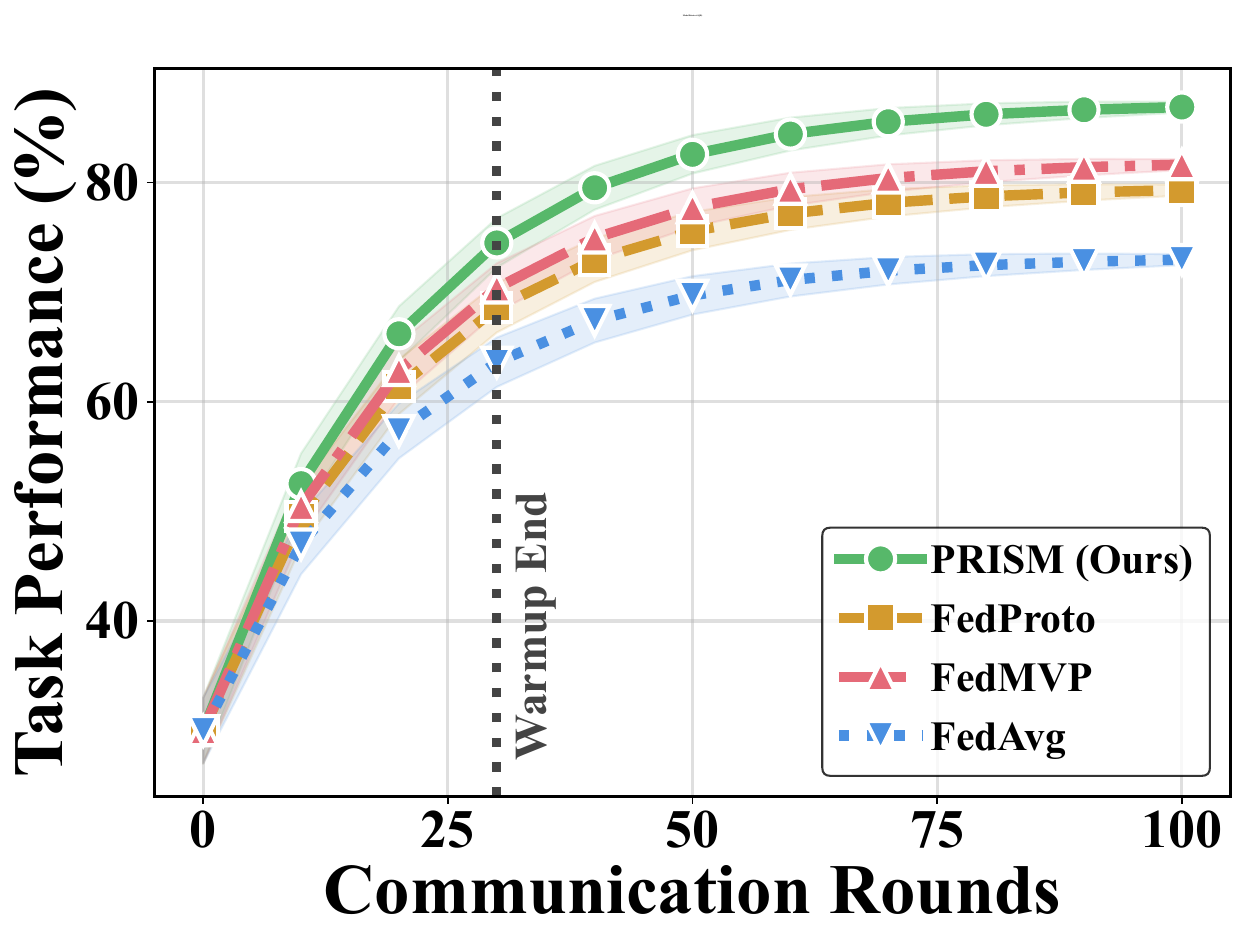}
        \label{fig:conv_qb}
    \end{minipage}
    \vspace{-20pt}
    \caption{\textbf{Convergence Analysis.} Solid lines and shaded areas denote mean and standard deviation over 5 trials.}
    \label{fig:convergence_all}
\end{figure}

\begin{figure}[!t]
    \centering
    
    \begin{minipage}[b]{0.3\linewidth}
        \centering
        {\tiny \hspace*{3pt} (a) Node Classification (Toys)} \\[1ex]
        \includegraphics[width=\linewidth]{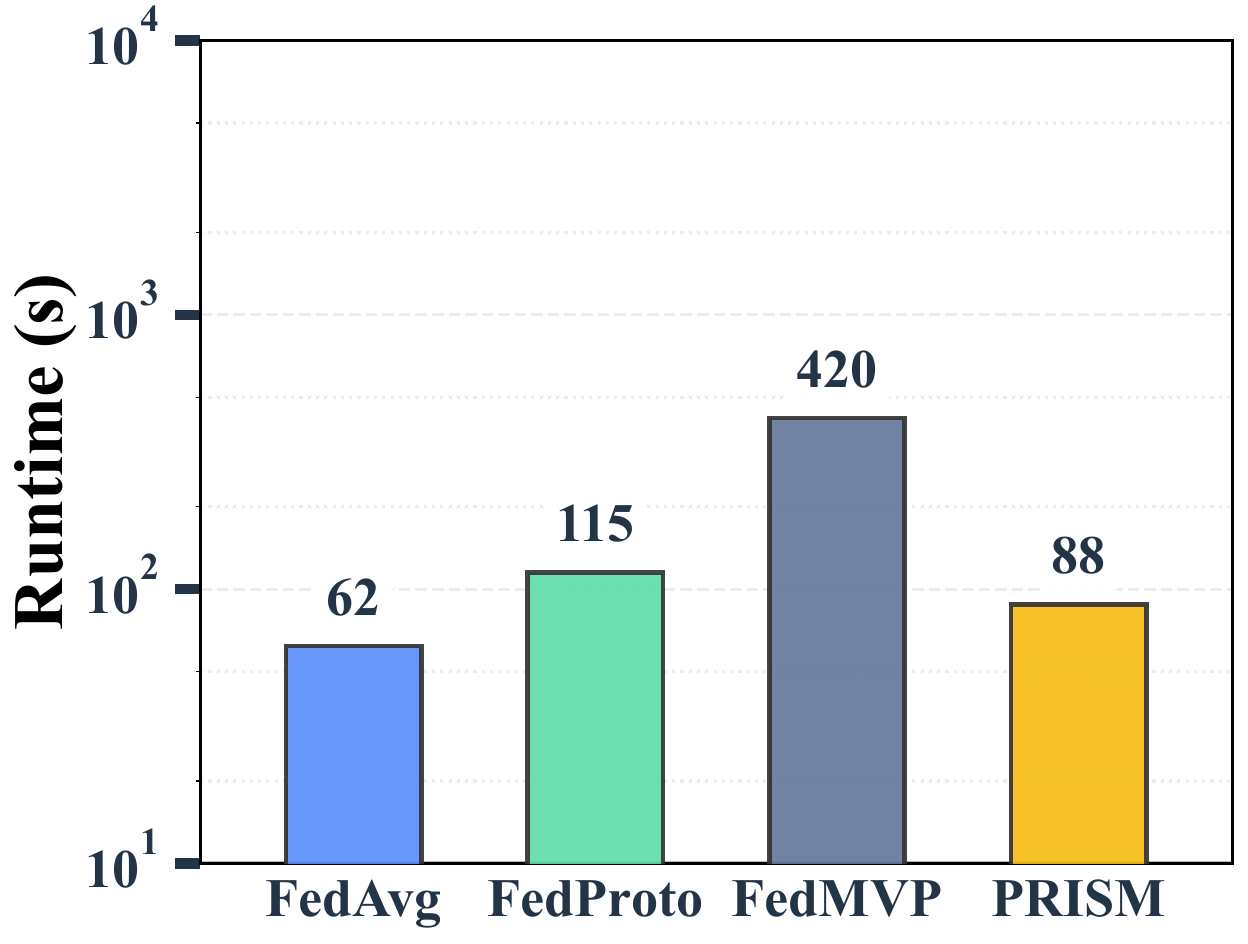}
        \label{fig:eff_node}
    \end{minipage}
    \hfill
    \begin{minipage}[b]{0.3\linewidth}
        \centering
        {\tiny \hspace*{3pt} (b) Modality Matching (KU)} \\[1ex]
        \includegraphics[width=\linewidth]{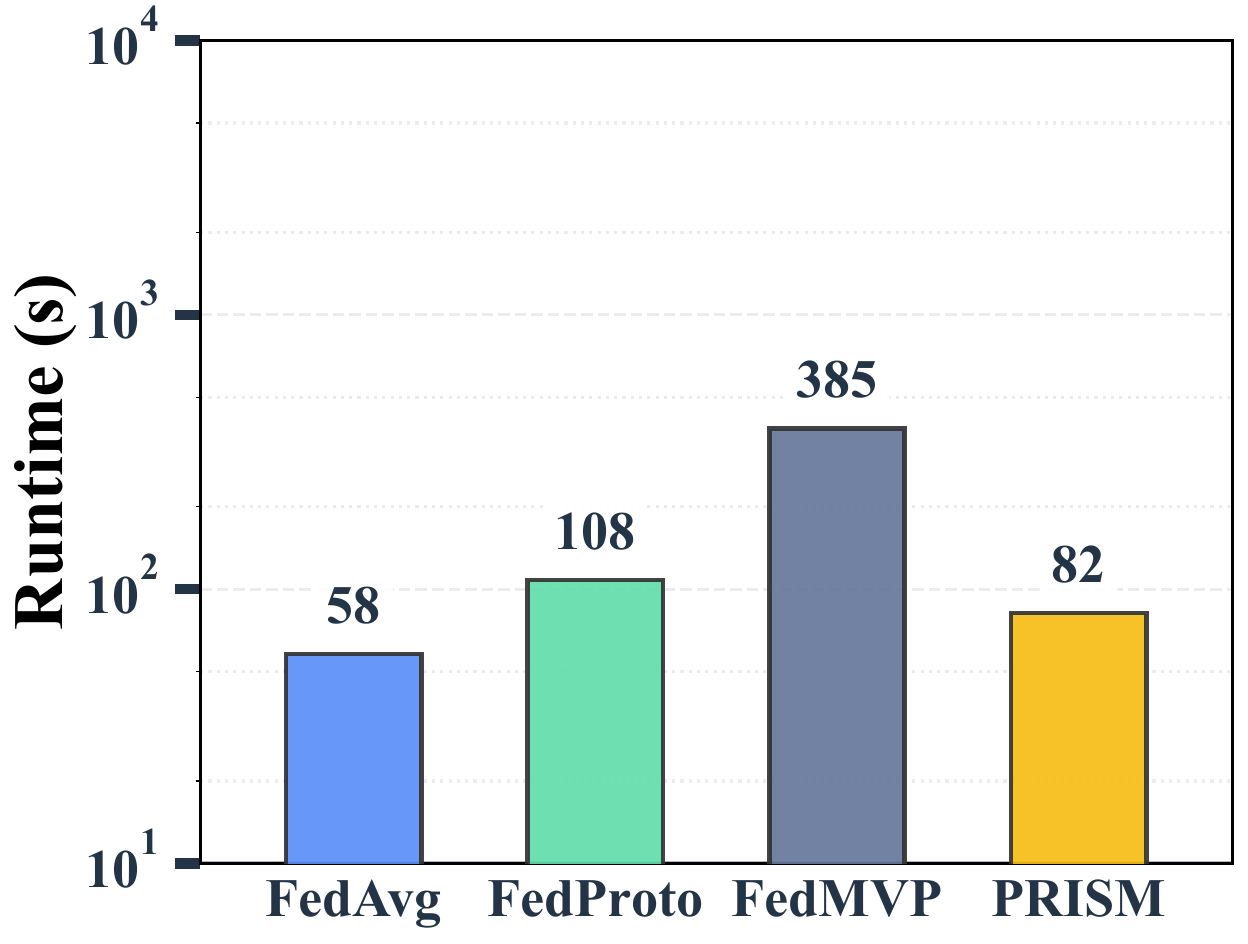}
        \label{fig:eff_matching}
    \end{minipage}
    \hfill
    \begin{minipage}[b]{0.3\linewidth}
        \centering
        {\tiny \hspace*{3pt} (c) Modal Retrieval (QB)} \\[1ex]
        \includegraphics[width=\linewidth]{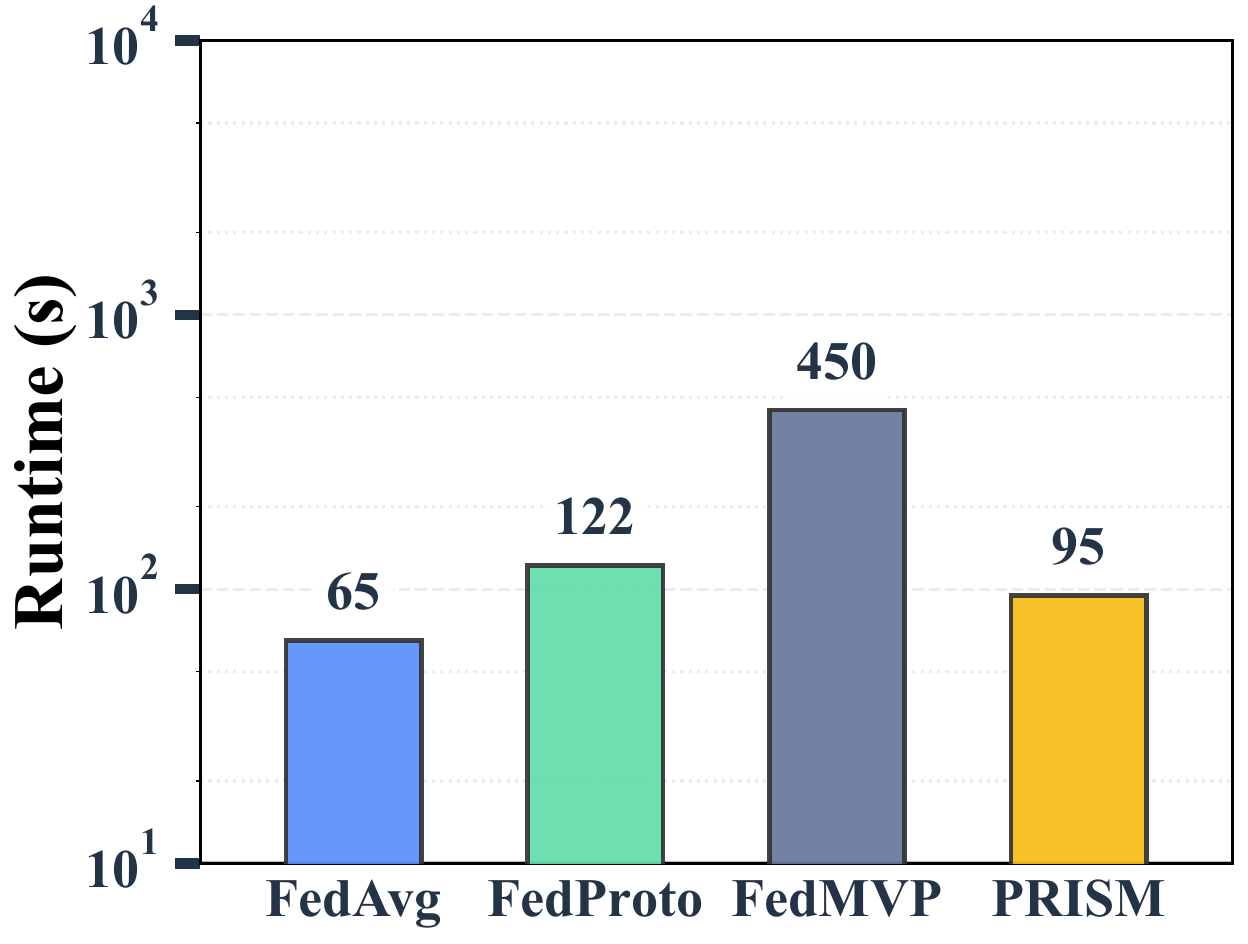}
        \label{fig:eff_retrieval}
    \end{minipage}
    \vspace{-20pt}
    \caption{\textbf{Efficiency Evaluation of PRISM.} These results compare PRISM's accuracy-efficiency trade-off with representative baselines.}
    \label{fig:efficiency}
\end{figure}

\begin{table*}[t]
    \centering
    \caption{Theoretical time, space, and communication complexity comparison of different methods.}
    \vspace{-4pt}
    \label{tab:complexity}
    \resizebox{\textwidth}{!}{
    \begin{tabular}{lccc}
        \toprule
        \textbf{Method} & \textbf{Time Complexity} & \textbf{Space Complexity} & \textbf{Comm. Complexity} \\
        \midrule
        FedAvg & $\mathcal{O}(REd(n+e))$& $\mathcal{O}(nd+e+|\Theta|)$& $\mathcal{O}(RN_C|\Theta|)$\\
        Fed-MHGAT & $\mathcal{O}(REd(n + e))$& $\mathcal{O}(nd+e+|\Theta|)$& $\mathcal{O}(RN_C|\Theta|)$\\
        FedProto & $\mathcal{O}(REd(n+e))$& $\mathcal{O}(nd+e+|\Theta|+|\mathcal{Y}|d)$& $\mathcal{O}(RN_C|\mathcal{Y}|d)$\\
        FedMAC & $\mathcal{O}(RE[d(n+e)+M^2dn b])$& $\mathcal{O}(nd+e+|\Theta|+M^2b^2+Mbd)$& $\mathcal{O}(RN_C|\Theta|)$\\
        FedC4 & $\mathcal{O}(T_{cond}d(n+e)+R(E+T_{GR})dk^2)$& $\mathcal{O}(nd+e+|\Theta|+kd + k^2)$& $\mathcal{O}( N_C^2kd+R(N_C|\Theta|+N_C \log N_C kd))$\\
        FedSPA & $\mathcal{O}(REd(n + e))$& $\mathcal{O}(nd+e+|\Theta|)$& $\mathcal{O}(RN_C|\Theta|)$\\
        \midrule
        \rowcolor[HTML]{F2F2F2} 
        \textbf{PRISM (Ours)} & \textbf{$\mathcal{O}(REd(n +e))$}& \textbf{$\mathcal{O}(nd+e+|\Theta|+J_bd+B_{prompt}d_q)$}& \textbf{$\mathcal{O}(RN_C(|\Theta|+J_rd + B_{prompt}d_q))$}\\
        \bottomrule
    \end{tabular}
    }

    \vspace{1.5ex}
    \small
    \raggedright
    \textit{Note}: $R$ is the number of communication rounds, $E$ is the number of local epochs per round, and $N_c$ is the number of clients. $n$ and $e$ denote the numbers of nodes and edges in a local client graph, respectively. $d$ represents the feature dimension, and $|\Theta|$ is the GNN backbone parameter size. $|\mathcal{Y}|$ is the number of classes. For FedMAC, $M$ is the number of modalities, $b$ is the contrastive-learning batch size. For FedC4, $k$ denotes the number of condensed synthetic nodes per client $(k\ll n)$, while $T_{\mathrm{cond}}$ and $T_{\mathrm{GR}}$ denote the numbers of graph condensation and graph rebuilding optimization steps, respectively. For PRISM, $d_q$ is the query dimension for retrieval, $J_b$ is the size of the global prototype bank, $J_r$ is the number of retrieved sparse prototypes per client per round, and $B_{\mathrm{prompt}}$ is the number of low-rank meta-prompt basis vectors. 
\end{table*}

To answer \textbf{EQ4}, we examine the convergence behavior and accuracy--efficiency trade-off of PRISM against representative baselines.
\label{subsec:communication_efficiency}

\begin{figure}[htb]
    \centering
    \includegraphics[width=\linewidth]{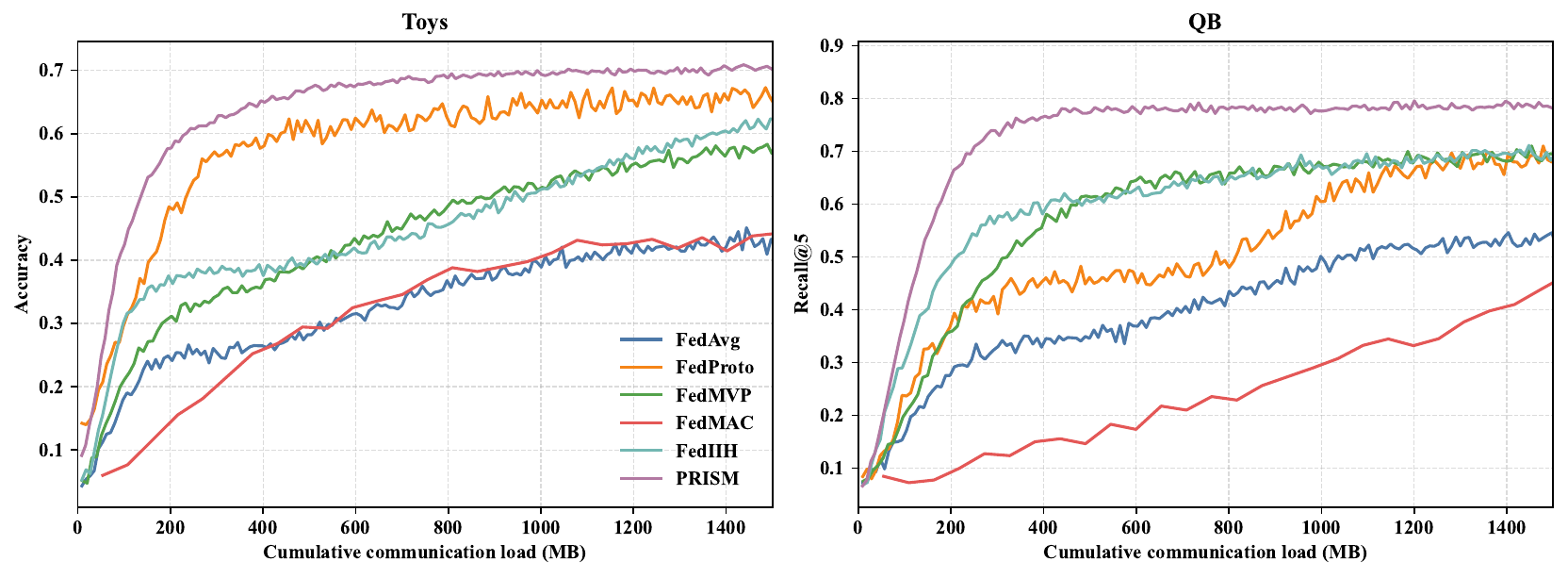}
    \vspace{-15pt}
    \caption{\textbf{Performance--communication trade-off under client-level modality deficiency.} We compare task performance as a function of cumulative communication load on Toys and QB.}
    \label{fig:comm_tradeoff}
\end{figure}

\textbf{Performance--communication trade-off.}
We first evaluate the cumulative communication load required to reach target performance on Toys and QB. 
As shown in Fig.~\ref{fig:comm_tradeoff}, PRISM achieves stronger performance with lower communication cost than representative FL, FGL, and missing-modality baselines. 
This advantage shows that sparse semantic retrieval offers a more efficient accuracy--communication trade-off than dense feature transfer or generic prototype aggregation, especially on QB where cross-modal retrieval depends more heavily on missing-modality semantics.

\begin{figure}[htb]
    \centering
    \includegraphics[width=\linewidth]{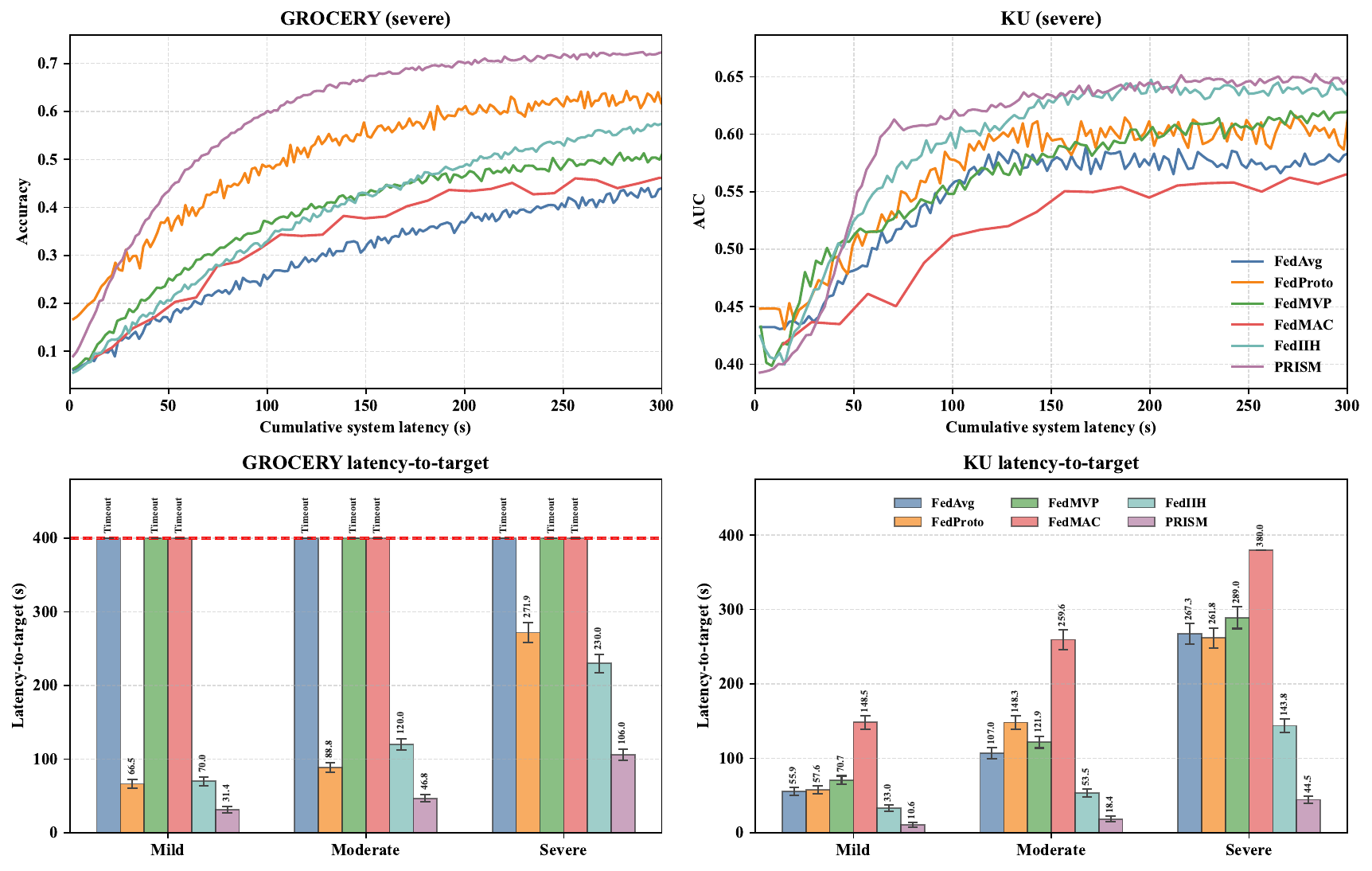}
    \vspace{-15pt}
    \caption{\textbf{Evaluation of system latency under heterogeneous wireless networks.} (Top) Task performance evolution (Accuracy on Grocery, AUC on KU) with respect to cumulative system latency under the severe bandwidth heterogeneity setting. (Bottom) The total system latency required to reach the target performance across mild, moderate, and severe network heterogeneity levels. Bars reaching the upper limit indicate the method failed to reach the target performance within the allocated budget. }
    \label{fig:wireless_latency}
\end{figure}

\textbf{Training Convergence.} Fig.~\ref{fig:convergence_all} compares training trajectories on three representative tasks. 
PRISM reaches higher steady-state performance and separates more clearly from baselines after the warm-up stage, when proactive retrieval is activated. 
This suggests that a stabilized prototype bank provides modality-deficient clients with more informative semantic anchors and improves optimization stability.

\textbf{Complexity Analysis.}
Table~\ref{tab:complexity} compares representative methods from different federated and multimodal graph learning paradigms. Most graph-based methods share the dominant GNN training cost $O(REd(n+e))$. PRISM maintains this order by using Proactive Cross-Modal Retrieval and Virtual Anchor Injection, without introducing high-order graph-size dependence. Compared with contrastive multimodal FL methods (e.g., FedMAC), PRISM avoids the $O(M^2dnb)$ batch-level contrastive cost; compared with graph condensation methods (e.g., FedC4), it avoids the extra condensation and graph rebuilding costs. Its additional overhead is limited to the Global Modality Bank, retrieved sparse prototypes, and Low-rank meta-prompt basis, leading to controllable space and communication costs.

\textbf{Efficiency Analysis.}
Empirically, Fig.~\ref{fig:efficiency} compares the accuracy and runtime trade-offs. Although PRISM introduces moderate computational overhead from sparse proactive retrieval and virtual anchor injection—which explains its slightly longer runtime compared to the vanilla FedAvg baseline (which only performs standard backbone training and parameter aggregation)—it consistently achieves a vastly superior accuracy-efficiency balance than other representative baselines. By avoiding dense cross-client feature transfer, $\mathcal{O}(n^2)$ attention computations, or expensive generative completion, PRISM remains highly practical and scalable for real-world multimodal federated graph learning.

\textbf{Communication Efficiency under Mobile Constraints.}
Empirically, Fig.~\ref{fig:comm_tradeoff} and Fig.~\ref{fig:wireless_latency} compare the communication costs and system latency trade-offs under bandwidth-limited and heterogeneous wireless conditions. Although PRISM transmits additional structural prompts and retrieved semantic values—which explains its slightly larger per-round payload compared to the vanilla FedAvg baseline (which only performs parameter aggregation)—it consistently achieves a vastly superior performance-communication balance than other representative baselines. By avoiding dense node-level missing-modality transfer, full prototype set synchronization, or heavy cross-modal alignment statistics, PRISM effectively mitigates the straggler bottlenecks induced by bandwidth-limited clients, remaining highly communication-efficient and practical for real-world mobile environments.
\section{Conclusion}

This paper studied multimodal federated graph learning under client-level modality deficiency, where a graph client may never observe one modality locally. We identify this deficiency as a graph-propagated semantic boundary error: the client starts message passing from an incomplete modality basis, and the resulting error is filtered by its topology. Motivated by this view, we proposed PRISM, a retrieve-globally-and-inject-structurally framework for federated cross-modal imputation. PRISM retrieves sparse multimodal prototypes from a server-side modality bank, instantiates them as virtual anchors, and uses structural meta-prompting to control their retrieval and injection according to the receiving graph. Diagnostic studies show that the same missing-modality error spreads differently across topologies and that topology-conditioned gating reduces ambiguous semantic contamination. Experiments across six multimodal graph benchmarks show consistent improvements on node classification, modality matching, and cross-modal retrieval, with clear gains under severe modality missingness. These results suggest that robust MM-FGL should treat missing modalities as a federated and topology-conditioned imputation problem rather than a purely local reconstruction problem.

\bibliographystyle{unsrt}
\bibliography{refs}

@article{blondel2008fast,
  title={Fast unfolding of communities in large networks},
  author={Blondel, Vincent D and Guillaume, Jean-Loup and Lambiotte, Renaud and Lefebvre, Etienne},
  journal={Journal of statistical mechanics: theory and experiment},
  volume={2008},
  number={10},
  pages={P10008},
  year={2008}
}

@article{kipf2016semi,
  title={Semi-supervised classification with graph convolutional networks},
  author={Kipf, Thomas N and Welling, Max},
  journal={arXiv preprint arXiv:1609.02907},
  year={2016}
}

@article{velivckovic2017graph,
  title={Graph attention networks},
  author={Veli{\v{c}}kovi{\'c}, Petar and Cucurull, Guillem and Casanova, Arantxa and Romero, Adriana and Lio, Pietro and Bengio, Yoshua},
  journal={arXiv preprint arXiv:1710.10903},
  year={2017}
}

@article{hamilton2017inductive,
  title={Inductive representation learning on large graphs},
  author={Hamilton, Will and Ying, Zhitao and Leskovec, Jure},
  journal={Advances in neural information processing systems},
  volume={30},
  year={2017}
}

@article{xu2018powerful,
  title={How powerful are graph neural networks?},
  author={Xu, Keyulu and Hu, Weihua and Leskovec, Jure and Jegelka, Stefanie},
  journal={arXiv preprint arXiv:1810.00826},
  year={2018}
}

@inproceedings{mcmahan2017communication,
  title={Communication-efficient learning of deep networks from decentralized data},
  author={McMahan, Brendan and Moore, Eider and Ramage, Daniel and Hampson, Seth and y Arcas, Blaise Aguera},
  booktitle={Artificial intelligence and statistics},
  pages={1273--1282},
  year={2017},
  organization={Pmlr}
}

@inproceedings{luanyuan2024mgnet,
  title={Mgnet: Learning correspondences via multiple graphs},
  author={Luanyuan, Dai and Du, Xiaoyu and Zhang, Hanwang and Tang, Jinhui},
  booktitle={Proceedings of the AAAI conference on Artificial Intelligence},
  volume={38},
  number={4},
  pages={3945--3953},
  year={2024}
}

@article{jia2023multimodal,
  title={Multimodal heterogeneous graph attention network},
  author={Jia, Xiangen and Jiang, Min and Dong, Yihong and Zhu, Feng and Lin, Haocai and Xin, Yu and Chen, Huahui},
  journal={Neural Computing and Applications},
  volume={35},
  number={4},
  pages={3357--3372},
  year={2023},
  publisher={Springer}
}

@article{aliakbari2025subgraph,
  title={Subgraph Federated Learning via Spectral Methods},
  author={Aliakbari, Javad and {\"O}stman, Johan and Panahi, Ashkan and others},
  journal={arXiv preprint arXiv:2510.25657},
  year={2025}
}

@article{tan2025s2fgl,
  title={S2FGL: Spatial Spectral Federated Graph Learning},
  author={Tan, Zihan and Huang, Suyuan and Wan, Guancheng and Huang, Wenke and Li, He and Ye, Mang},
  journal={arXiv preprint arXiv:2507.02409},
  year={2025}
}

@inproceedings{tan2022fedproto,
  title={Fedproto: Federated prototype learning across heterogeneous clients},
  author={Tan, Yue and Long, Guodong and Liu, Lu and Zhou, Tianyi and Lu, Qinghua and Jiang, Jing and Zhang, Chengqi},
  booktitle={Proceedings of the AAAI conference on artificial intelligence},
  volume={36},
  number={8},
  pages={8432--8440},
  year={2022}
}

@inproceedings{singha2025fedmvp,
  title={FedMVP: Federated Multimodal Visual Prompt Tuning for Vision-Language Models},
  author={Singha, Mainak and Roy, Subhankar and Mehrotra, Sarthak and Jha, Ankit and Abdar, Moloud and Banerjee, Biplab and Ricci, Elisa},
  booktitle={Proceedings of the IEEE/CVF International Conference on Computer Vision},
  pages={17869--17878},
  year={2025}
}

@inproceedings{nguyen2024fedmac,
  title={Fedmac: Tackling partial-modality missing in federated learning with cross-modal aggregation and contrastive regularization},
  author={Nguyen, Manh Duong and Nguyen, Trung Thanh and Pham, Huy Hieu and Hoang, Trong Nghia and Le Nguyen, Phi and Huynh, Thanh Trung},
  booktitle={2024 22nd International Symposium on Network Computing and Applications (NCA)},
  pages={278--285},
  year={2024},
  organization={IEEE}
}

@inproceedings{xie2024mh,
  title={Mh-pflgb: Model heterogeneous personalized federated learning via global bypass for medical image analysis},
  author={Xie, Luyuan and Lin, Manqing and Xu, ChenMing and Luan, Tianyu and Zeng, Zhipeng and Qian, Wenjun and Li, Cong and Fang, Yuejian and Shen, Qingni and Wu, Zhonghai},
  booktitle={International Conference on Medical Image Computing and Computer-Assisted Intervention},
  pages={534--545},
  year={2024},
  organization={Springer}
}

@article{zhang2024ninerec,
  title={Ninerec: A benchmark dataset suite for evaluating transferable recommendation},
  author={Zhang, Jiaqi and Cheng, Yu and Ni, Yongxin and Pan, Yunzhu and Yuan, Zheng and Fu, Junchen and Li, Youhua and Wang, Jie and Yuan, Fajie},
  journal={IEEE Transactions on Pattern Analysis and Machine Intelligence},
  year={2024},
  publisher={IEEE}
}

@inproceedings{yan2025graph,
  title={When graph meets multimodal: benchmarking and meditating on multimodal attributed graph learning},
  author={Yan, Hao and Li, Chaozhuo and Yin, Jun and Yu, Zhigang and Han, Weihao and Li, Mingzheng and Zeng, Zhengxin and Sun, Hao and Wang, Senzhang},
  booktitle={Proceedings of the 31st ACM SIGKDD Conference on Knowledge Discovery and Data Mining V. 2},
  pages={5842--5853},
  year={2025}
}

@inproceedings{radford2021learning,
  title={Learning transferable visual models from natural language supervision},
  author={Radford, Alec and Kim, Jong Wook and Hallacy, Chris and Ramesh, Aditya and Goh, Gabriel and Agarwal, Sandhini and Sastry, Girish and Askell, Amanda and Mishkin, Pamela and Clark, Jack and others},
  booktitle={International conference on machine learning},
  pages={8748--8763},
  year={2021},
  organization={PmLR}
}

@article{dosovitskiy2020image,
  title={An image is worth 16x16 words: Transformers for image recognition at scale},
  author={Dosovitskiy, Alexey and Beyer, Lucas and Kolesnikov, Alexander and Weissenborn, Dirk and Zhai, Xiaohua and Unterthiner, Thomas and Dehghani, Mostafa and Minderer, Matthias and Heigold, Georg and Gelly, Sylvain and others},
  journal={arXiv preprint arXiv:2010.11929},
  year={2020}
}

@article{he2021fedgraphnn,
  title={Fedgraphnn: A federated learning system and benchmark for graph neural networks},
  author={He, Chaoyang and Balasubramanian, Keshav and Ceyani, Emir and Yang, Carl and Xie, Han and Sun, Lichao and He, Lifang and Yang, Liangwei and Yu, Philip S and Rong, Yu and others},
  journal={arXiv preprint arXiv:2104.07145},
  year={2021}
}

@article{li2026mmopenfgl,
  title={MM-OpenFGL: A Comprehensive Benchmark for Multimodal Federated Graph Learning},
  author={Li, Xunkai and Ai, Yuming and Zhu, Yinlin and Lu, Haodong and Zhang, Yi and Fu, Guohao and Fan, Bowen and Dai, Qiangqiang and Li, Rong-Hua and Wang, Guoren},
  journal={arXiv preprint arXiv:2601.22416},
  year={2026}
}

@article{li2024openfgl,
  title={Openfgl: A comprehensive benchmark for federated graph learning},
  author={Li, Xunkai and Zhu, Yinlin and Pang, Boyang and Yan, Guochen and Yan, Yeyu and Li, Zening and Wu, Zhengyu and Zhang, Wentao and Li, Rong-Hua and Wang, Guoren},
  journal={arXiv preprint arXiv:2408.16288},
  year={2024}
}

@inproceedings{hou2022graphmae,
  title={GraphMAE: Self-Supervised Masked Graph Autoencoders},
  author={Hou, Zhenyu and Liu, Xiao and Cen, Yukuo and Dong, Yuxiao and Yang, Hongxia and Wang, Chunjie and Tang, Jie},
  booktitle={Proceedings of the 28th ACM SIGKDD Conference on Knowledge Discovery and Data Mining},
  pages={594--604},
  year={2022},
  doi={10.1145/3534678.3539321}
}

@inproceedings{zhang2021subgraph,
  title={Subgraph Federated Learning with Missing Neighbor Generation},
  author={Zhang, Ke and Yang, Carl and Li, Xiaoxiao and Sun, Lichao and Yiu, Siu Ming},
  booktitle={Advances in Neural Information Processing Systems},
  volume={34},
  pages={6671--6682},
  year={2021}
}

@inproceedings{chen2025rethinking,
  title={Rethinking Client-oriented Federated Graph Learning},
  author={Chen, Zekai and Li, Xunkai and Zhu, Yinlin and Li, Rong-Hua and Wang, Guoren},
  booktitle={Proceedings of the 34th ACM International Conference on Information and Knowledge Management},
  pages={393--402},
  year={2025}
}

@article{zhang2024deep,
  title={Deep efficient private neighbor generation for subgraph federated learning},
  author={Zhang, Ke and Sun, Lichao and Ding, Bolin and Yiu, Siu Ming and Yang, Carl},
  journal={arXiv preprint arXiv:2401.04336},
  year={2024}
}

@inproceedings{tan2025fedspa,
  title={FedSPA: Generalizable Federated Graph Learning under Homophily Heterogeneity},
  author={Tan, Zihan and Wan, Guancheng and Huang, Wenke and Li, He and Zhang, Guibin and Yang, Carl and Ye, Mang},
  booktitle={Proceedings of the Computer Vision and Pattern Recognition Conference},
  pages={15464--15475},
  year={2025}
}

@inproceedings{yu2025modeling,
  title={Modeling inter-intra heterogeneity for graph federated learning},
  author={Yu, Wentao and Chen, Shuo and Tong, Yongxin and Gu, Tianlong and Gong, Chen},
  booktitle={Proceedings of the AAAI Conference on Artificial Intelligence},
  volume={39},
  number={21},
  pages={22236--22244},
  year={2025}
}

@inproceedings{pietilainen2009mobiclique,
  title     = {MobiClique: Middleware for Mobile Social Networking},
  author    = {Pietil{\"a}inen, Anna-Kaisa and Oliver, Earl and LeBrun, Jason and Varghese, George and Diot, Christophe},
  booktitle = {Proceedings of the 2nd ACM Workshop on Online Social Networks},
  pages     = {49--54},
  year      = {2009}
}

@inproceedings{miluzzo2008sensing,
  title     = {Sensing Meets Mobile Social Networks: The Design, Implementation and Evaluation of the CenceMe Application},
  author    = {Miluzzo, Emiliano and Lane, Nicholas D. and Fodor, Krist{\'o}f and Peterson, Ronald and Lu, Hong and Musolesi, Mirco and Eisenman, Shane B. and Zheng, Xiao and Campbell, Andrew T.},
  booktitle = {Proceedings of the 6th ACM Conference on Embedded Networked Sensor Systems},
  pages     = {337--350},
  year      = {2008},
  doi       = {10.1145/1460412.1460445}
}

@article{dong2023gnniot,
  title   = {Graph Neural Networks in IoT: A Survey},
  author  = {Dong, Guimin and Tang, Mingyue and Wang, Zhiyuan and Gao, Jiechao and Guo, Sikun and Cai, Lihua and Gutierrez, Robert and Campbell, Bradford and Barnes, Laura E. and Boukhechba, Mehdi},
  journal = {ACM Transactions on Sensor Networks},
  volume  = {19},
  number  = {2},
  pages   = {1--50},
  year    = {2023},
  doi     = {10.1145/3565973}
}

@inproceedings{li2018dcrnn,
  title     = {Diffusion Convolutional Recurrent Neural Network: Data-Driven Traffic Forecasting},
  author    = {Li, Yaguang and Yu, Rose and Shahabi, Cyrus and Liu, Yan},
  booktitle = {International Conference on Learning Representations},
  year      = {2018}
}

@inproceedings{xu2022v2xvit,
  title     = {V2X-ViT: Vehicle-to-Everything Cooperative Perception with Vision Transformer},
  author    = {Xu, Runsheng and Xiang, Hao and Tu, Zhengzhong and Xia, Xin and Yang, Ming-Hsuan and Ma, Jiaqi},
  booktitle = {European Conference on Computer Vision},
  pages     = {107--124},
  year      = {2022},
  doi       = {10.1007/978-3-031-19842-7_7}
}

@inproceedings{hu2022where2comm,
  title     = {Where2comm: Communication-Efficient Collaborative Perception via Spatial Confidence Maps},
  author    = {Hu, Yue and Fang, Shaoheng and Lei, Zixing and Zhong, Yiqi and Chen, Siheng},
  booktitle = {Advances in Neural Information Processing Systems},
  year      = {2022}
}

@article{wang2023mhagnn,
  title   = {MhaGNN: A Novel Framework for Wearable Sensor-Based Human Activity Recognition Combining Multi-Head Attention and Graph Neural Networks},
  author  = {Wang, Yan and Wang, Xin and Yang, Hongmei and others},
  journal = {IEEE Transactions on Instrumentation and Measurement},
  volume  = {72},
  pages   = {1--14},
  year    = {2023},
  doi     = {10.1109/TIM.2023.3276004}
}

\vspace{-1.3cm}

\begin{IEEEbiography}
[{\includegraphics[width=0.8in,height=1in,clip,keepaspectratio]{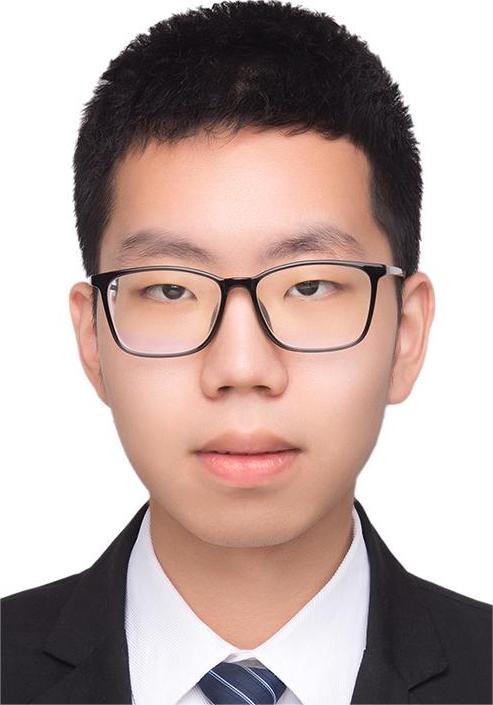}}]{Zekai Chen}
is currently pursuing his Master's degree in Computer Science at Beijing Institute of Technology under the supervision of Professor Rong-Hua Li. He obtained his Bachelor degree in Computer Science from the same institution in 2024. His research focuses on Graph Machine Learning and AI for Science (AI4Science). 
\end{IEEEbiography}

\vspace{-1.6cm}

\begin{IEEEbiography}
[{\includegraphics[width=0.8in,height=1in,clip,keepaspectratio]{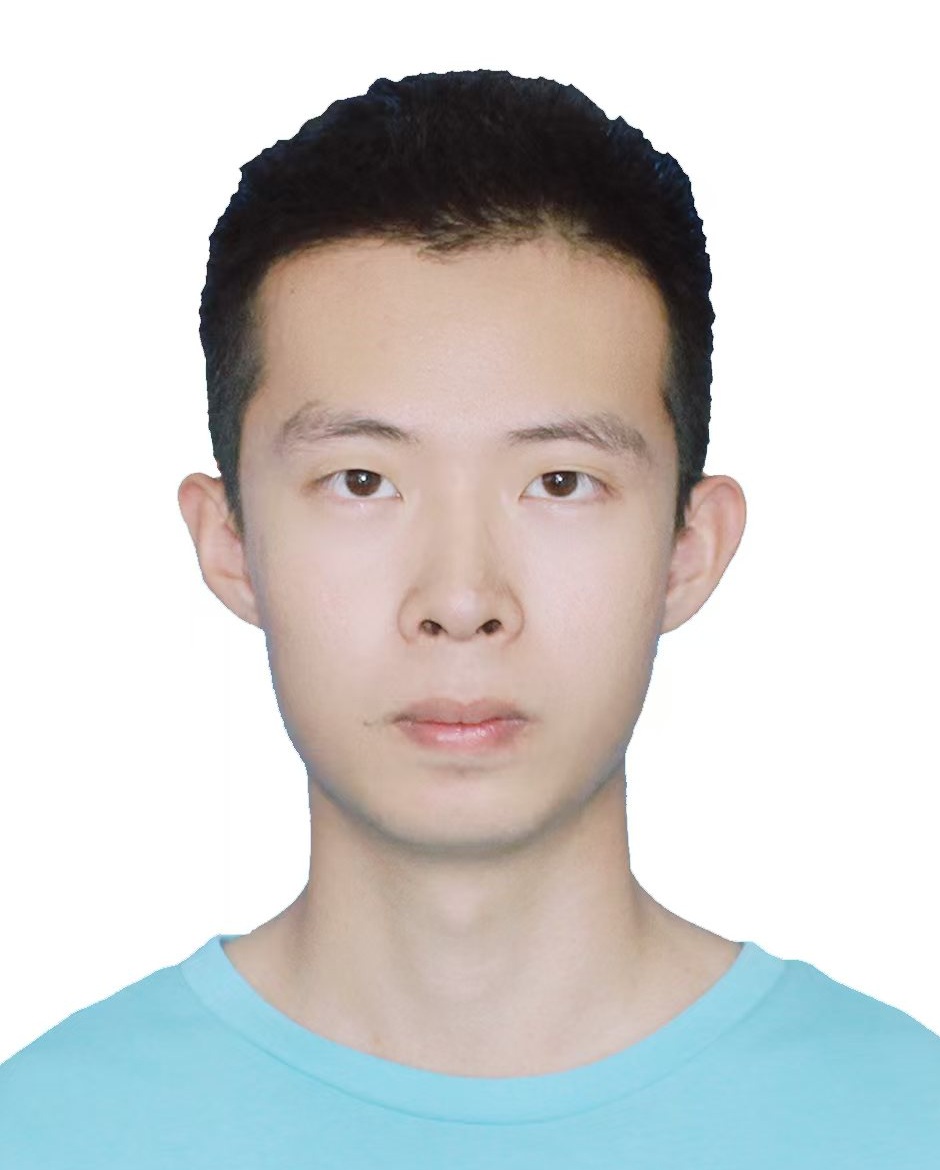}}]{Miao Zhang}
is currently pursuing the B.S. degree in Data Science and Big Data Technology at the School of Computer Science, Beijing Institute of Technology, Beijing, China. His research interests include Graph Machine Learning.
\end{IEEEbiography}

\vspace{-1.6cm}

\begin{IEEEbiography}
[{\includegraphics[width=0.8in,height=1in,clip,keepaspectratio]{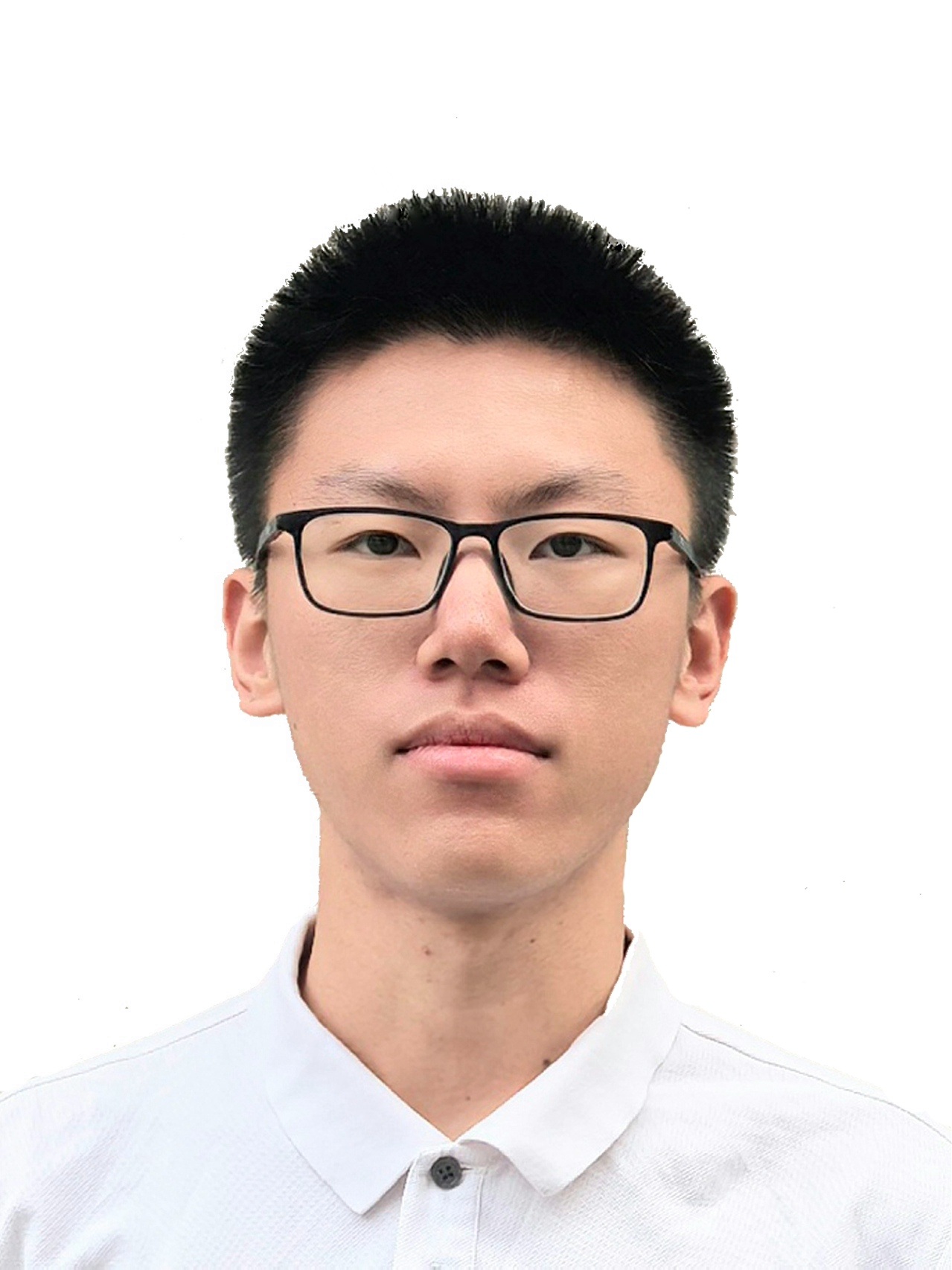}}]{Jiayang Xing}
is currently pursuing the B.S. degree in Computer Science and Technology at the School of Computer Science, Beijing Institute of Technology, Beijing, China. His research interests include Graph Machine Learning.
\end{IEEEbiography}

\vspace{-1.6cm}

\begin{IEEEbiography}
[{\includegraphics[width=0.8in,height=1in,clip,keepaspectratio]{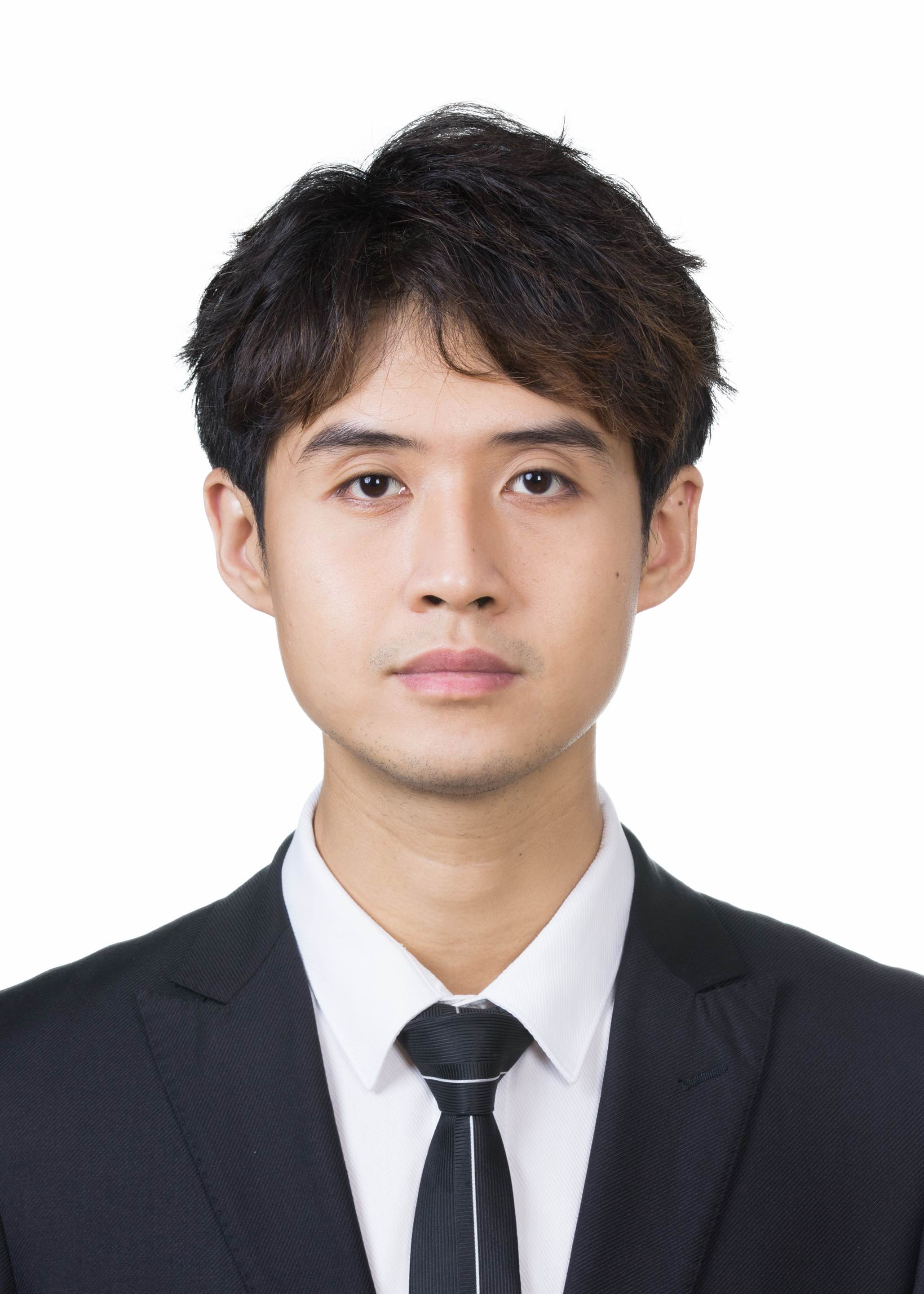}}]{Xunkai Li}
is currently pursuing the PhD degree in Beijing Institute of Technology, advised by Prof. Rong-Hua Li. He received the BS degree from Shandong University in 2022. His research interest lies in Data-centric Graph Intelligence (Data-centric AI, Graph Machine Learning, and AI4Science). He has published 10+ papers in top ML/DB/DM/AI conferences such as ICML, VLDB, WWW, AAAI.
\end{IEEEbiography}

\vspace{-1.6cm}

\begin{IEEEbiography}
[{\includegraphics[width=0.8in,height=1in,clip,keepaspectratio]{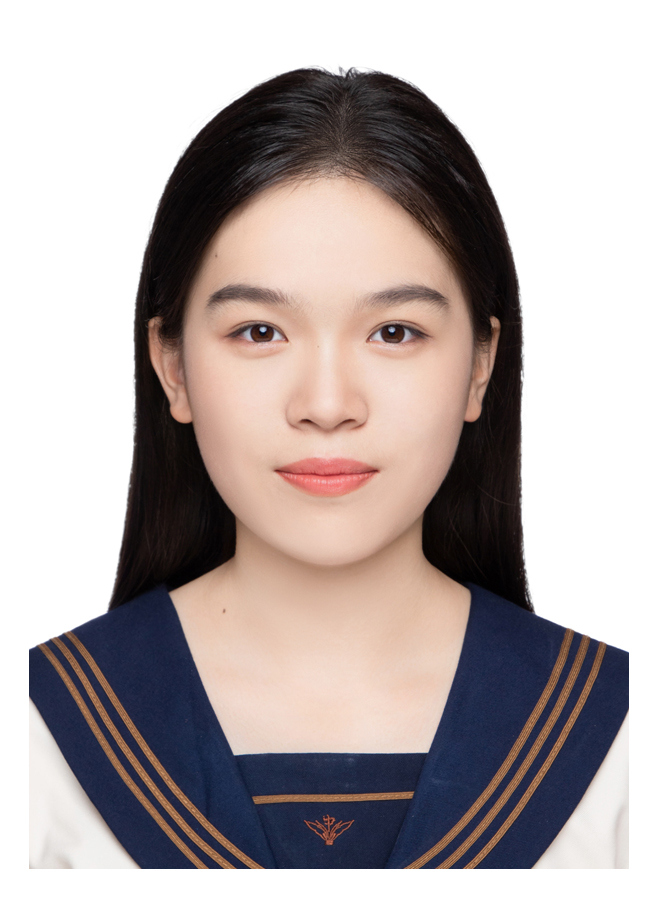}}]{Xun Wu} is currently pursuing the B.S. degree in data science and big data technology with the School of Computer Science, Beijing Institute of Technology, Beijing, China. Her research interest lies in Graph Machine Learning.
\end{IEEEbiography}

\vspace{-1.6cm}

\begin{IEEEbiography}[{\includegraphics[width=0.8in,height=1in,clip,keepaspectratio]{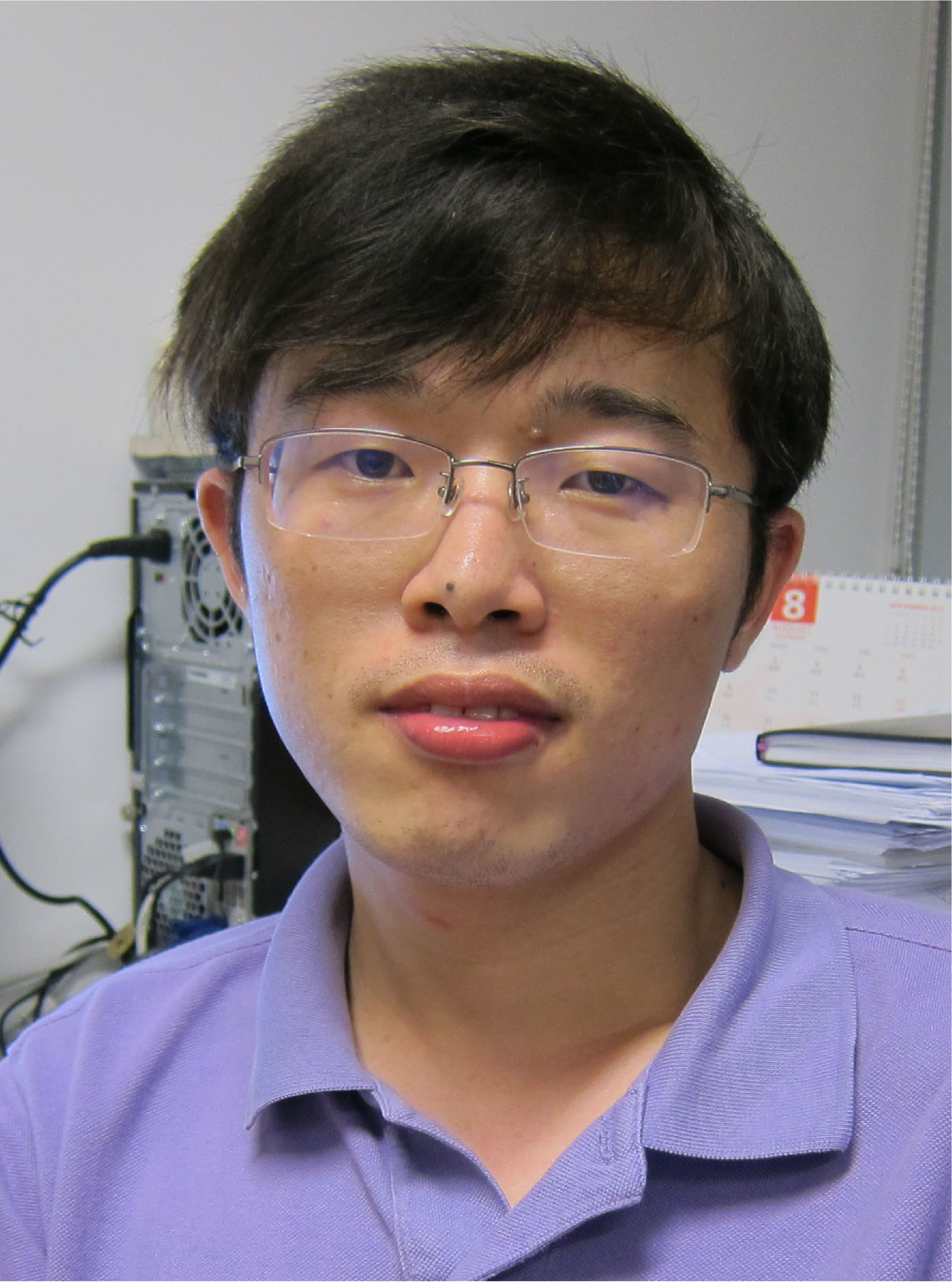}}]{Rong-Hua Li}
 received the PhD degree from the Chinese University of Hong Kong, in 2013. He is currently a professor with the Beijing Institute of Technology (BIT), Beijing, China. His research interests include graph data management and mining, social network analysis, graph computational systems, and graph-based machine learning.
\end{IEEEbiography}

\vspace{-1.6cm}

\begin{IEEEbiography}[{\includegraphics[width=0.8in,height=1in,clip,keepaspectratio]{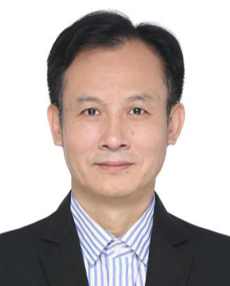}}]{Guoren Wang}
 received the BS, MS, and PhD degrees from the Department of Computer Science, Northeastern University, China, in 1988, 1991, and 1996, respectively. Currently, he is a professor with the Beijing Institute of Technology (BIT), Beijing, China. His research interests include graph data management, graph mining, and graph computational systems.
\end{IEEEbiography}

\vfill

\end{document}